\documentclass{article}

\usepackage{microtype}
\usepackage{graphicx}
\usepackage{subcaption}
\usepackage{booktabs} 

\usepackage{hyperref}

\usepackage[accepted]{icml2026}

\usepackage{amsmath}
\usepackage{amssymb}
\usepackage{mathtools}
\usepackage{amsthm}

\usepackage{multirow}

\usepackage[capitalize,noabbrev]{cleveref}

\theoremstyle{plain}

\theoremstyle{definition}

\theoremstyle{remark}

\usepackage[textsize=tiny]{todonotes}

\icmltitlerunning{CoGenCast: A Coupled Autoregressive–Flow Generative Framework for  Time Series Forecasting}

\begin{document}

\twocolumn[
  \icmltitle{CoGenCast: A Coupled Autoregressive-Flow Generative Framework \\for   Time Series Forecasting}



  \icmlsetsymbol{equal}{*}

  \begin{icmlauthorlist}
    \icmlauthor{Mingyue Cheng}{yyy}
    \icmlauthor{Yaguo Liu}{yyy}
    \icmlauthor{Daoyu Wang}{yyy}
    \icmlauthor{Xiaoyu Tao}{yyy}
    \icmlauthor{Qi Liu*}{yyy}

  \end{icmlauthorlist}

  \icmlaffiliation{yyy}{
State Key Laboratory of Cognitive Intelligence, University of
Science and Technology of China, Hefei, China}

  \icmlcorrespondingauthor{Qi Liu}{ qiliuql@ustc.edu.cn}

  \icmlkeywords{Machine Learning, ICML}

  \vskip 0.3in
]



\printAffiliationsAndNotice{}  
\begin{abstract}
Time series forecasting can be  viewed as a generative problem that requires both semantic understanding over contextual conditions and stochastic modeling of continuous temporal dynamics. Existing approaches typically rely on either autoregressive large language models (LLMs) for semantic context modeling or diffusion-like models for continuous probabilistic generation. However, neither method alone can adequately model both aspects simultaneously. In this work, we propose CoGenCast, a hybrid generative framework that couples pre-trained LLMs with flow-matching mechanism for effective time series forecasting. Specifically, we reconfigure  pre-trained decoder-only LLMs into a  native forecasting encoder–decoder backbone by modifying only the attention topology, enabling bidirectional context encoding and causal representation generation. Building on this, a flow-matching mechanism  is further integrated to model temporal evolution, capturing continuous stochastic dynamics conditioned on the autoregressively generated representation. Notably, CoGenCast naturally supports multimodal forecasting and cross-domain unified training. Extensive experiments on multiple benchmarks show that CoGenCast achieves  competitive performance compared to previous  baselines. Code is available at \url{https://github.com/liuyaguo/_CoGenCast}.
\end{abstract}

\section{Introduction}

Time series forecasting supports a wide range of real-world decision-making processes, including energy~\cite{zhou2024sdwpf}, finance~\cite{feng2019temporal} and healthcare~\cite{qiu2024tfb}. The core philosophy of  forecasting~\cite{liu2023adaptive} is to predict future values conditioned on historical observations, while future evolution is often full of  complexity and uncertainty.  To handle this, previous works typically implement a simple regression mapping~\cite{nie2022time}. However, recently, an increasing number of methods  employ more powerful mapping functions, such as LLMs~\cite{niu2025langtime,jiang2025fstllm} and diffusion-like~\cite{guo2025dynamical} models. These methods reformulate forecasting as a conditional generative problem~\cite{liu2024generative}.

Current generative forecasting methods can be divided into two main categories: LLM-based methods and diffusion-like models. First, LLM-based methods leverage the strong  semantic understanding and generation capabilities of LLMs. These can be further branched   into  tuning-based methods~\cite{jin2024time,xie2025chatts},  which finetune LLMs  to align them with numerical time series modeling, and training-free methods~\cite{liu2024lstprompt,cheng2026can}, which keep LLMs frozen and rely on prompting or reasoning mechanisms to perform forecasting. Second, diffusion-like models focus on modeling the probabilistic evolution of future time series in continuous-valued spaces. These approaches effectively capture uncertainty  in future temporal trajectories. Representative works such as NsDiff~\cite{ye2025non} and TSFlow~\cite{kollovieh2025flow} learn continuous stochastic temporal dynamics, enabling high-quality probabilistic generation~\cite{tashiro2021csdi}.

Although the above generative forecasting methods achieve promising results, we argue that an ideal forecasting approach should simultaneously possess dual capabilities: semantic understanding over contextual conditions and stochastic modeling of continuous temporal dynamics. Taking electricity load forecasting~\cite{wang2025timedart} as a concrete example, a model should  achieve a deep  understanding of diverse contextual conditions, such as historical load patterns, urban population, and weather variables, to capture the underlying semantic dependencies. Meanwhile, future electricity demand is closely  linked to unpredictable future weather shifts, holiday effects, and other uncertain factors~\cite{rasul2021autoregressive, zhang2024trajectory} requiring  stochastic  modeling of continuous temporal dynamics.

Building upon the analysis above, we propose CoGenCast, a hybrid generative framework that couples pre-trained LLMs with flow-matching mechanism for effective time series forecasting. Specifically, we introduce a  reconfiguration of  pre-trained decoder-only LLMs into a native forecasting encoder–decoder backbone. Instead of retraining LLMs from scratch, we modify only the attention topology to construct a bidirectional encoder for comprehensive understanding over contextual conditions and a causal decoder for autoregressive representation generation. A flow-matching denoising decoder is further integrated to model temporal evolution, effectively capturing continuous stochastic dynamics conditioned on the autoregressively generated representation. Notably, unlike conventional flow-matching methods, our approach targets the mean velocity field, enabling efficient one-step generation with  low latency. As a result, CoGenCast naturally supports multimodal forecasting and cross-domain unified training.  Extensive experiments on multiple benchmarks show that CoGenCast achieves  competitive performance compared to previous  baselines.

We summarize our main contributions as follows:
\begin{itemize}
    \item We  reconfigure  pre-trained decoder-only LLMs into a native forecasting encoder-decoder backbone by modifying only the attention topology.
    \item We propose CoGenCast, a hybrid generative framework that couples pre-trained LLMs with continuous flow-matching mechanism for  time series forecasting.
    \item We conduct extensive experiments on multiple benchmarks, demonstrating that CoGenCast achieves  competitive performance compared to previous  baselines.
\end{itemize}

\section{Related Work}

\subsection{LLM-based Time Series Forecasting}
Recently, LLM-based~\cite{jin2024position} methods  become increasingly prevalent in time series forecasting  due to the strong semantic understanding~\cite{pan2024s} and generation~\cite{bian2024multi} capabilities of LLMs. Existing LLM-based time series forecasting methods can be broadly categorized into tuning-based and training-free methods. Within the tuning-based category, PromptCast~\cite{xue2023promptcast} reformulates forecasting as a sentence-to-sentence task using  text prompts; Time-LLM~\cite{jin2024time} utilizes a learnable interface to reprogram continuous patches into the linguistic space; LLM4TS~\cite{chang2025llm4ts} combines the LLM’s feature extraction capabilities with a specialized linear head for numerical output; TokenCast~\cite{tao2025values} implements symbolic discretization to map series into a unified vocabulary for autoregressive generation; ChatTS~\cite{xie2025chatts} aligns time series as a standalone modality with LLMs via synthetic data to enable open-ended reasoning; and Time-R1~\cite{luo2025time} feeds the entire sequence as a direct reasoning input while employing reinforcement learning to optimize explicit slow-thinking chains. Conversely, within the training-free category, TimeReasoner~\cite{cheng2026can} effectively induces multi-step temporal reasoning capabilities via multi-step prompting strategies.
\begin{figure*}[htbp]
    \centering
    \includegraphics[width=\textwidth]{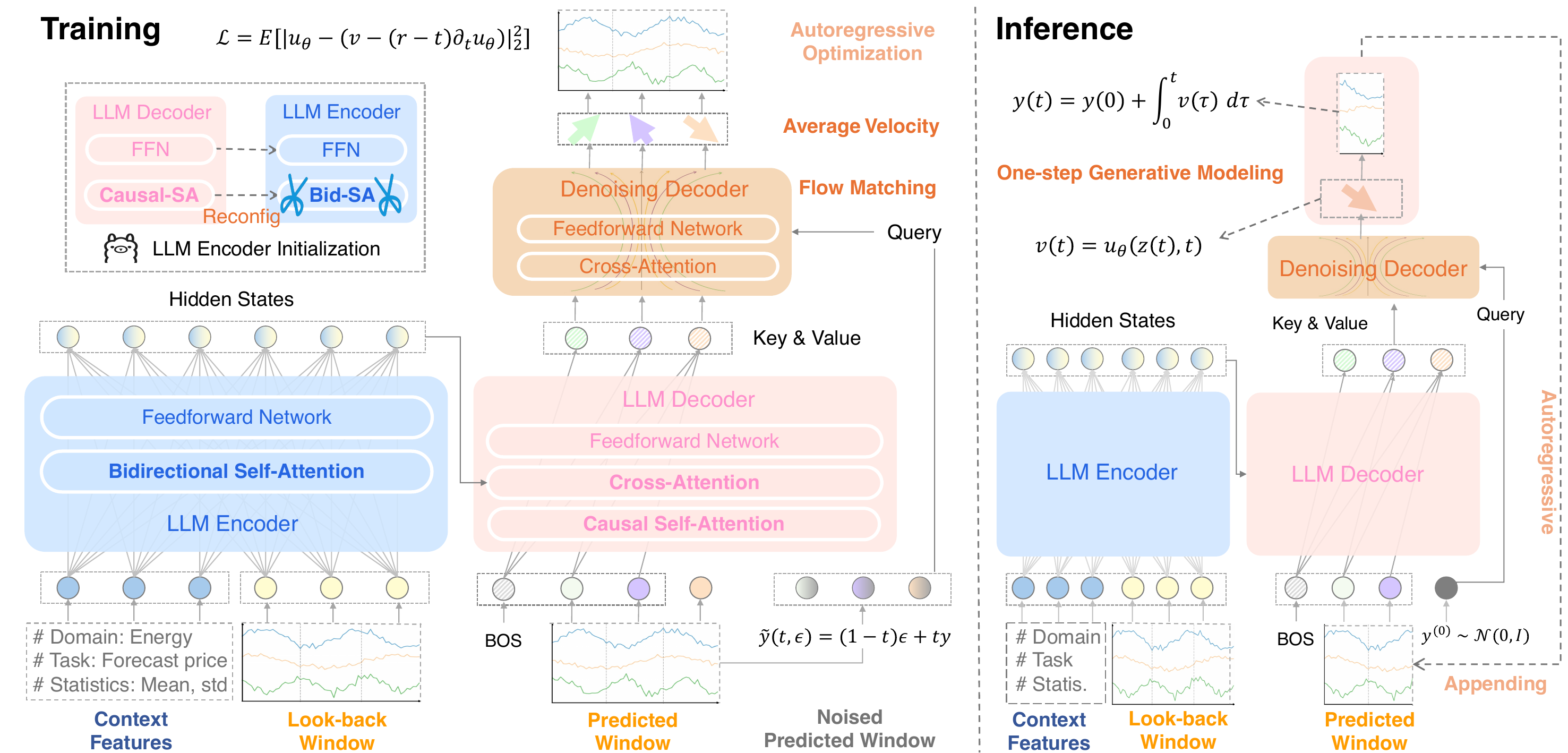}
    \caption{Overview of our proposed CoGenCast.
\textbf{Left (Training):} We reconfigure decoder-only LLMs into an encoder--decoder backbone by attention-only modification, and perform
continuous flow-matching mechanism conditioned on the LLM-generated representation.
\textbf{Right (Inference):} Future patches are generated autoregressively and sampled via one-step flow-matching
generation with low-latency.}

    \label{fig:framework}
\end{figure*}

\subsection{Diffusion-like Time Series Forecasting}
Diffusion models  become superior generative tools for time series forecasting due to their  powerful capabilities of effectively capturing temporal uncertainty~\cite{wu2025k2vae}. Early research primarily focused on foundational architectural adaptations: TimeGrad~\cite{rasul2021autoregressive} pioneers the integration of an autoregressive framework with denoising diffusion models, while CSDI~\cite{tashiro2021csdi} utilizes a non-autoregressive score-based model to capture spatio-temporal correlations. Subsequent studies shift toward condition-enhanced mechanisms to improve effectiveness: TimeDiff~\cite{shen2023non} introduces future mixup and autoregressive initialization to strengthen pattern capture; NsDiff~\cite{ye2025non} incorporates a location-scale noise model  for adaptive noise scheduling in non-stationary data; and DyDiff~\cite{guo2025dynamical} improves  forecasting performance by modeling temporal transitions within the diffusion process. To bridge the efficiency gap of iterative diffusion models, flow-matching models have emerged as its faster generation. FlowTS~\cite{hu2024flowts} accelerates generation by learning straight-line transport paths via rectified flow, while TSFlow~\cite{kollovieh2025flow} incorporates Gaussian process priors to  align these trajectories with temporal dynamics.   Additionally, FreqFlow~\cite{moghadas2025freqflow} shifts the process into the frequency domain to capture long-term patterns. Despite  these  methods make significant progress, no single approach  adequately satisfies the dual requirements of    semantic understanding and continuous stochastic modeling. To address this, we propose CoGenCast, a hybrid generative framework that couples pre-trained LLMs with continuous flow-matching mechanism for   time series forecasting to  model both aspects simultaneously.

\section{CoGenCast}
In this section, we provide  the formal problem definition and present an overview of our proposed CoGenCast. We also detail the implementation process.

\subsection{Problem Formulation}
We consider the  time series forecasting problem where the goal is  to forecast future sequences based on historical context features and numerical sequences~\cite{cheng2025cross}. Let $\mathbf{X}_{\text{hist}} = \{x_1, \dots, x_L\} \in \mathbb{R}^{L \times D}$ denote the look-back window of length $L$ with $D$ variables. The objective is to forecast the predicted window $\mathbf{X}_{\text{pred}} = \{x_{L+1}, \dots, x_{L+H}\} \in \mathbb{R}^{H \times D}$ of length $H$. Additionally,  let $\mathbf{C}$  denote the context features (e.g., domain knowledge, task instruction,statistics information). The problem~\cite{cheng2025comprehensive} is formulated as learning the conditional probability distribution $P(\mathbf{X}_{\text{pred}} \mid \mathbf{X}_{\text{hist}}, \mathbf{C})$.

\subsection{Framework Overview}
We propose a hybrid generative framework to model the conditional probability distribution
$P(\mathbf{X}_{\text{pred}} \mid \mathbf{X}_{\text{hist}}, \mathbf{C})$ by coupling pre-trained LLMs with flow-matching mechanism.
We first reconfigure  pre-trained decoder-only LLMs into a native forecasting encoder-decoder backbone by modifying only the attention topology.
Specifically, we initialize the LLM encoder by reusing the pre-trained decoder parameters while replacing causal self-attention with bidirectional self-attention. The LLM decoder keeps causal self-attention to preserve temporal causality, and introduces a newly added cross-attention  that queries the encoder hidden states as keys and values. To capture stochastic continuous dynamics with  the causal representation generation, we incorporate a continuous flow-matching mechanism conditioned on the LLM-generated representation. Specifically, we define a probability path  and train a flow-matching denoising decoder to predict the corresponding temporal velocity field. Moreover, by learning an interval-conditioned average velocity, the temporal evolution can approximate the entire
transport process within a single step, which enables efficient sampling without iterative denoising. During inference, each patch is produced autoregressively and sampled efficiently via one-step generation with low-latency.

\subsection{Coupled Training}

\paragraph{Autoregressive Language Encoder-Decoder.}
We first train the backbone to perform autoregressive encoder--decoder generation,  producing semantically grounded and causal representations that serve as conditions for subsequent continuous flow-matching mechanism. Unlike conventional decoder-only LLMs~\cite{zhang2026star,wang2025paperarena,wang2026steppo}, which rely solely on causal attention, time series forecasting  requires both  bidirectional understanding and causal generation. To meet these two requirements, we reconfigure  decoder-only LLMs into a native forecasting encoder-decoder backbone by modifying only the attention topology: the encoder adopts bidirectional self-attention to fuse the look-back window and context features~\cite{zhang2025alphacast}, while the decoder preserves causal self-attention to generate future representations autoregressively. Given context features $\mathbf{C}$ , we obtain text embeddings $\mathbf{h}^{\text{enc\_in}}_{\text{text}} \in \mathbb{R}^{M \times D_{\text{model}}}$ using the LLM's native tokenizer and embedding layer. In parallel, the look-back window $\mathbf{X}_{\text{hist}} \in \mathbb{R}^{L \times D}$ is partitioned into $M=L/P$ non-overlapping patches $\mathbf{x}_{1:M}$, and projected into the same latent dimension $D_{\text{model}}$ via a learnable linear layer to yield historical patch embeddings $\mathbf{h}^{\text{enc\_in}}_{1:M}$. We then concatenate the text and patch embeddings along the sequence~\cite{yang2013travel} dimension and feed them into the bidirectional encoder to obtain fused contextual states $\mathbf{h}^{\text{enc\_out}}$:
\begin{equation}
    \mathbf{h}^{\text{enc\_out}} = \text{Encoder}\!\left(\text{Concat}\!\left(\mathbf{h}^{\text{enc\_in}}_{\text{text}}, \mathbf{h}^{\text{enc\_in}}_{1:M}\right)\right).
\end{equation}

To generate future structures in a forecasting-native manner, we similarly segment the predicted window $\mathbf{Y} \in \mathbb{R}^{H \times D}$ into $N=H/P$ patches $\mathbf{y}_{1:N}$ and map them into latent embeddings $\mathbf{z}_{1:N}$.  To align with autoregressive training and ensure strict causality, we employ a shifted-input strategy by prepending a learnable begin-of-sequence (BOS) embedding and excluding the final patch embedding $\mathbf{z}^{\text{dec\_in}}_{1:N}$:
\begin{equation}
    \mathbf{z}^{\text{dec\_in}}_{1:N} = \text{Concat}(\text{BOS}, \mathbf{z}_{1:N-1}).
\end{equation}
The shifted embeddings~\cite{ye2020few} are then fed into the causal decoder, which combines (i) causal self-attention to model temporal dependencies without look-ahead and (ii) cross-attention over $\mathbf{h}^{\text{enc\_out}}$ to incorporate multimodal historical context. The  autoregressive  states $\mathbf{z}^{\text{dec\_out}}_{1:N}$ are:
\begin{equation}
    \mathbf{z}^{\text{dec\_out}}_{1:N} = \text{Decoder}(\mathbf{z}^{\text{dec\_in}}_{1:N}, \mathbf{h}^{\text{enc\_out}}),
\end{equation}

where each $\mathbf{z}^{\text{dec\_out}}_{j}$ is computed under a strict causal mask and  summarizes information only from $\mathbf{y}_{1:j-1}$ together with the encoded history, making it a valid autoregressive representation for step-$j$ forecasting.  These hidden states  condition the subsequent  continuous flow-matching mechanism, enabling to jointly model  semantic understanding~\cite{tao2026cast} and continuous stochastic dynamics adequately.

\paragraph{Continuous Flow-Matching Mechanism.}
In contrast to standard diffusion~\cite{wang2025diffuse, chen2025deconstructing} or flow-matching methods~\cite{geng2025improved, geng2026mean,lu2025bidirectional,lu2026one} that estimate an instantaneous vector field at a single time point, our formulation explicitly models the average velocity over a finite time interval. This interval-level velocity directly characterizes the displacement between two time steps, thereby reducing the error induced by local instantaneous velocity estimation and coarse numerical integration.  To construct a probability path that transports samples from a source distribution to the target  distribution, we adopt a direct linear interpolation between the clean future patches $\mathbf{y}_{1:N}$ and Gaussian noise $\boldsymbol{\epsilon} \sim \mathcal{N}(0, \mathbf{I})$. For any time $t \in [0,1]$, the intermediate noisy state $\hat{\mathbf{y}}_{1:N}$ is defined as a convex combination of signal and noise:
\begin{equation}
    \hat{\mathbf{y}}_{1:N} = (1 - t)\boldsymbol{\epsilon} + t\mathbf{y}_{1:N},
\end{equation}
 where $t=0$ corresponds to pure noise, while $t=1$ recovers the clean patches. Geometrically, this linear interpolation induces a straight-line trajectory in data~\cite{miao2024unified} space, characterized by a constant base velocity direction $\mathbf{v}$:
\begin{equation}
    \mathbf{v} = \mathbf{y}_{1:N} - \boldsymbol{\epsilon}.
\end{equation}
This constant-velocity property substantially simplifies the underlying flow dynamics and provides a principled basis for high-quality few-step (and even one-step) generative modeling. We first project the interpolated noisy patches into the latent space using a linear projection layer, yielding the noised embeddings $\hat{\mathbf{z}}^{\text{in}}_{1:N}$. The denoising decoder  strictly enforces the autoregressive constraint through an  alignment strategy: the noised embeddings $\hat{\mathbf{z}}^{\text{in}}_{1:N}$ are used as queries, while the conditional hidden states $\mathbf{z}^{\text{dec\_out}}_{1:N}$ are used as keys and values. Notably, due to the shifted-input mechanism in the backbone, each context state $\mathbf{z}^{\text{dec\_out}}_{j}$ has already aggregated information from preceding clean patches $\mathbf{y}_{1:j-1}$. Unlike conventional flow matching, our estimator is conditioned not only on the current time $t$ but also on a target step $r$ (with $r>t$), enabling the model to explicitly perceive the integration interval length. The velocity prediction~\cite{gardner1985exponential,hyndman2008automatic,taylor2018forecasting} for the $j$-th patch is  formulated as:
\begin{equation}
    \mathbf{u}^{\text{out}}_j = \text{DenoisingDecoder}(\hat{\mathbf{z}}^{\text{in}}_j, t, r, \mathbf{z}^{\text{dec\_out}}_{1:j}),
\end{equation}
where the denoising decoder serves as a velocity estimator.  This formulation ensures that the predicted velocity $\mathbf{u}^{\text{out}}_j$ depends only on the noised embedding $\hat{\mathbf{z}}^{\text{in}}_j$, the interval $[t,r]$, and the available autoregressive context $\mathbf{z}^{\text{dec\_out}}_{1:j}$, thereby preventing any information leakage from future positions.

\paragraph{Optimization Objective.}
To explicitly guide the model to learn the average velocity along the temporal trajectory, we employ a Jacobian-Vector Product (JVP) corrected optimization objective. During training, we sample the current time step $t \sim \mathcal{U}[0, 1]$ and a target time step $r \sim \mathcal{U}[t, 1]$. To approximate the true average velocity required to span the interval $[t, r]$, we apply a first-order Taylor expansion correction using the gradient of the velocity field. The final objective minimizes the mean squared error (MSE) between the predicted velocity ${\mathbf{u}}$ and the JVP-corrected target. Formally, the loss function is expressed as:
\begin{equation}
     \mathbb{E}_{t, r, \boldsymbol{\epsilon}, \mathbf{y}} \left[ \frac{1}{N} \sum_{j=1}^{N} \left\| \mathbf{u}^{\text{out}}_j - \left( \mathbf{v}_{ j} - (r - t) \frac{\partial \mathbf{u}^{\text{out}}_j}{\partial t} \right) \right\|^2_2 \right],
\end{equation}
where $\frac{\partial \mathbf{u}^{\text{out}}_j}{\partial t}$ denotes the time partial derivative of the velocity field, which is efficiently computed via JVP. Minimizing this objective explicitly penalizes velocity variations, thereby compelling the model to learn straight temporal trajectories.

\subsection{One-step Inference}

\paragraph{Autoregressive Patch Generation.}
As illustrated in Figure~\ref{fig:framework} (Right), inference proceeds iteratively in an autoregressive manner. For generating the $j$-th patch, the previously generated patches $\hat{\mathbf{y}}_{1:j-1}$ are first appended to the predicted window, which initiates with the BOS token. This sequential appending serves to update the conditional keys and values used by the decoder, effectively capturing the causal dependencies between continuous temporal intervals. Unlike traditional point-regression models, our autoregressive appending maintains global semantic coherence by leveraging the hidden states of previously generated patches as contextual conditions for the subsequent generation step.

\paragraph{One-step Generative Modeling.}
Within each autoregressive step, we initiate the generation of the $j$-th patch by sampling a pure Gaussian noise patch $\mathbf{y}^{(0)}_j \sim \mathcal{N}(0, \mathbf{I})$ (labeled as $\mathbf{y}^{(0)}$ in Figure~\ref{fig:framework}). This noise patch is projected into the latent space via a linear layer to obtain the  noisy representation $\mathbf{z}^{(0)}_j$.  Given the current noisy representation and the available autoregressive context, the denoising decoder functions as a velocity estimator and predicts the interval-conditioned patch-level velocity field. Notably, by learning an interval-conditioned average velocity, the transport process can approximate the entire transformation from noise ($t=0$) to clean prediction ($t=1$) within a single step, which enables efficient sampling without iterative denoising. The output $\mathbf{y}^{\text{out}}_j$ is then recovered by integrating the predicted velocity over the time interval as shown in Eq.~(8):
\begin{equation}
    \mathbf{y}^{\text{out}}_j = \mathbf{y}^{(0)}_j + \int_{0}^{1} \mathbf{u}(\mathbf{z}_{\tau}, \tau, \mathbf{z}^{\text{dec\_out}}_{1:j}) \, d\tau .
\end{equation}
Finally, the integrated result serves as the generated patch $\hat{\mathbf{y}}_j$ and is appended to the predicted window to form $\hat{\mathbf{y}}_{1:j}$, which will be used as the input context for the next autoregressive inference step. Notably, this one-step generation, corresponding to a single function evaluation,  reduces  inference latency while maintaining high-performance forecasting.

\section{Experiments}
In this section, we conduct comprehensive experiments to evaluate the performance of our proposed CoGenCast on  multiple benchmark datasets. Additionally, we perform extensive ablation studies and exploration analysis.

\subsection{Experimental Setup}
\paragraph{Datasets.} We conduct extensive experiments on publicly available datasets, specifically including Energy~\cite{liu2024time}, ETT (4 subsets)~\cite{zhou2021informer}, Environment (abbreviated as Environ.)~\cite{liu2024time}, Exchange~\cite{wu2021autoformer}, Health~\cite{liu2024time},   Wind~\cite{li2022generative}, and Solar~\cite{lai2018modeling}. Detailed statistics are summarized in Table~\ref{tab:datasets}. To clearly illustrate the practical scenarios and scale of these datasets, the table provides an overview of each dataset’s domain, the number of samples, and the number of variables per sample. A more comprehensive description can be found in Appendix~\ref{sec:dataset description}.

\paragraph{Baselines.} To ensure a comprehensive evaluation, we compare our method with thirteen representative baselines spanning three major categories. 
First, for LLM-based methods, we select TokenCast~\cite{tao2025values}, LLM4TS~\cite{chang2025llm4ts} and Time-LLM~\cite{jin2024time}. TokenCast adopts an autoregressive next-token prediction, LLM4TS adapts pre-trained LLMs using two-stage fine-tuning, and Time-LLM reprograms frozen LLMs. Second, we select generative-based methods such as FlowTS~\cite{hu2024flowts},  CDPM~\cite{zhang2025conditional} and TimeGrad~\cite{rasul2021autoregressive}. FlowTS constructs a continuous probability path using flow-matching, CDPM employs conditional diffusion probabilistic models, and TimeGrad adopts the integration of an autoregressive framework with denoising diffusion models. Third, we compare with  deep-based methods, including ConvTimeNet~\cite{cheng2025convtimenet}, TimeDART~\cite{wang2025timedart}, iTransformer~\cite{liu2024itransformer}, TimesNet~\cite{wu2022timesnet}, PatchTST~\cite{nie2022time}, DLinear~\cite{zeng2023transformers} and Autoformer~\cite{wu2021autoformer}. Further details are provided in Appendix~\ref{app:baselines}.

\begin{table}[t]
\centering
\caption{Overview of the dataset statistics, including domain, the number of samples, and the number of variables per sample.}
\label{tab:datasets}
\small
\setlength{\tabcolsep}{4pt}
\renewcommand{\arraystretch}{1.05}
\resizebox{\columnwidth}{!}{%
\begin{tabular}{cccc}
\toprule
\textsc{Datasets} & \textsc{Domain} & \textsc{Variables} & \textsc{Length} \\
\midrule
Energy & Energy & 1 & 1,479 \\
ETTh1/ETTh2 & Energy & 7 & 17,420 \\
ETTm1/ETTm2 & Energy & 7 & 69,680 \\
Environment & Environment & 1 & 15,248 \\
Exchange & Finance & 8 & 7,588 \\
Health & Health & 1 & 1,389 \\
Wind & Electricity & 7 & 48,673 \\
Solar & Electricity & 137 & 52,560 \\
\bottomrule
\end{tabular}%
}
\end{table}

\begin{table*}[!t]
\centering
\caption{Time series forecasting results. All results are averaged Mean Squared
Error (MSE) and Mean Absolute Error(MAE) from the same look-back window of $L=96$ and 4 different predicted windows of \{12, 24, 36, 48\}. The best results are in \textbf{bold} and the second best are \underline{underlined}. Full results are detailed in Appendix~\ref{sec:full result}.}
\label{tab:mainresult}
\setlength{\tabcolsep}{1.1pt}
\renewcommand{\arraystretch}{1.2}
\small
\resizebox{\textwidth}{!}{%
\begin{tabular}{l|cc|cccccc|cccccc|cccccccccccccc}
\toprule

\multirow{2}{*}{\textsc{Methods}}
& \multicolumn{2}{c|}{\textsc{Ours}}
& \multicolumn{6}{c|}{\textsc{LLM-based}}
& \multicolumn{6}{c|}{\textsc{Generative-based}}
& \multicolumn{14}{c}{\textsc{Deep-based}} \\[2pt]

& \multicolumn{2}{c|}{\textbf{CoGenCast}}
& \multicolumn{2}{c}{TokenCast}
& \multicolumn{2}{c}{Time-LLM}
& \multicolumn{2}{c|}{LLM4TS}
& \multicolumn{2}{c}{FlowTS}
& \multicolumn{2}{c}{CDPM}
& \multicolumn{2}{c|}{TimeGrad}
& \multicolumn{2}{c}{ConvTimeNet}
& \multicolumn{2}{c}{TimeDART}
& \multicolumn{2}{c}{iTransformer}
& \multicolumn{2}{c}{TimesNet}
& \multicolumn{2}{c}{PatchTST}
& \multicolumn{2}{c}{DLinear}
& \multicolumn{2}{c}{Autoformer} \\[2pt]

\textsc{Metrics}
& MSE & MAE
& MSE & MAE & MSE & MAE & MSE & MAE
& MSE & MAE & MSE & MAE & MSE & MAE
& MSE & MAE & MSE & MAE & MSE & MAE
& MSE & MAE & MSE & MAE & MSE & MAE & MSE & MAE \\
\midrule

Energy
& \textbf{0.293} & \textbf{0.395}
& \underline{0.316} & \underline{0.412} & 0.318 & 0.413 & 0.325 & 0.424
& 0.418 & 0.473 & 0.373 & 0.447 & 0.674 & 0.634
& 0.317 & 0.416 & 0.323 & 0.421 & 0.326 & 0.427 & 0.330 & 0.427 & 0.327 & 0.424 & 0.335 & 0.431 & 0.506 & 0.542 \\

ETTh1
& \textbf{0.328} & \textbf{0.367}
& \underline{0.345} & \underline{0.381} & 0.364 & 0.391 & 0.348 & 0.382
& 0.381 & 0.413 & 0.350 & 0.383 & 0.893 & 0.776
& 0.351 & 0.382 & 0.354 & 0.384 & 0.357 & 0.388 & 0.361 & 0.389 & 0.367 & 0.390 & 0.363 & 0.388 & 0.495 & 0.469 \\

ETTh2
& 0.159 & \textbf{0.246}
& 0.191 & 0.278 & 0.206 & 0.291 & 0.195 & 0.281
& 0.218 & 0.300 & 0.193 & 0.279 & 0.571 & 0.541
& 0.165 & 0.266 & \underline{0.145} & 0.253 & 0.194 & 0.281 & 0.164 & 0.266 & \textbf{0.142} & \underline{0.251} & 0.165 & 0.266 & 0.192 & 0.301 \\

ETTm1
& \textbf{0.224} & \textbf{0.285}
& 0.248 & 0.310 & 0.304 & 0.349 & 0.253 & 0.314
& 0.327 & 0.374 & 0.302 & 0.352 & 0.801 & 0.614
& \underline{0.240} & \underline{0.305} & \underline{0.240} & 0.309 & 0.244 & 0.309 & 0.243 & 0.311 & 0.243 & 0.312 & 0.247 & 0.315 & 0.375 & 0.432 \\

ETTm2
& \textbf{0.109} & \textbf{0.200}
& 0.141 & 0.247 & 0.141 & 0.247 & 0.145 & 0.252
& 0.197 & 0.305 & 0.191 & 0.300 & 0.710 & 0.510
& \underline{0.119} & \underline{0.226} & 0.122 & 0.232 & 0.124 & 0.231 & 0.127 & 0.236 & 0.126 & 0.235 & 0.122 & 0.230 & 0.238 & 0.365 \\

Environ.
& \textbf{0.301} & \textbf{0.381}
& 0.316 & 0.396 & 0.320 & 0.400 & 0.316 & \underline{0.395}
& 0.349 & 0.425 & 0.334 & 0.419 & 0.844 & 0.680
& 0.311 & 0.397 & 0.314 & 0.399 & 0.315 & 0.403 & 0.314 & 0.401 & \underline{0.309} & 0.397 & 0.319 & 0.406 & 0.366 & 0.461 \\

Exchange
& \textbf{0.031} & \textbf{0.118}
& 0.037 & 0.134 & 0.035 & 0.130 & 0.037 & 0.133
& 0.061 & 0.185 & 0.046 & 0.153 & 0.320 & 0.444
& \underline{0.033} & 0.125 & 0.034 & 0.127 & 0.034 & 0.126 & 0.040 & 0.138 & \underline{0.033} & 0.127 & \underline{0.033} & \underline{0.123} & 0.087 & 0.214 \\

Health
& \textbf{1.444} & \textbf{0.812}
& 1.576 & \underline{0.848} & 1.611 & 0.887 & 1.679 & 0.923
& 1.736 & 0.943 & 1.591 & 0.897 & 2.486 & 1.439
& 1.535 & 0.857 & \underline{1.510} & 0.852 & 1.592 & 0.889 & 1.690 & 0.896 & 1.540 & 0.857 & 1.685 & 0.903 & 1.828 & 1.018 \\

Wind
& \textbf{0.531} & \textbf{0.425}
& 0.544 & 0.444 & 0.543 & 0.444 & 0.591 & 0.484
& 0.651 & 0.497 & 0.583 & 0.468 & 0.954 & 0.666
& 0.544 & 0.443 & 0.541 & 0.443 & \underline{0.537} & \underline{0.432} & 0.541 & 0.442 & 0.539 & 0.441 & 0.546 & 0.446 & 0.830 & 0.609 \\

Solar
& \textbf{0.224} & \textbf{0.289}
& 0.233 & 0.298 & 0.233 & 0.298 & 0.239 & 0.307
& 0.369 & 0.422 & 0.350 & 0.410 & 0.911 & 0.764
& 0.229 & 0.296 & 0.236 & 0.301 & \underline{0.225} & \underline{0.291} & 0.235 & 0.301 & 0.232 & 0.298 & 0.240 & 0.305 & 0.661 & 0.626 \\

\bottomrule
\end{tabular}
}
\end{table*}

\paragraph{Implementation Details.} In the experiments, for all datasets, we set the look-back window length to $L = 96$  and the predicted window length with $H \in \{12, 24, 36, 48\}$. We report two widely used forecasting metrics:
mean squared error (MSE) and mean absolute error (MAE). Our backbone architecture is initialized with the pre-trained
Qwen3-0.6B model parameters for both the encoder and decoder. For the continuous flow-matching mechanism, we adopt a linear noise scheduler, which naturally matches our straight-trajectory formulation and enables
efficient one-step sampling at inference time. More comprehensive implementation details can be found in Appendix~\ref{app:implementation}.

\subsection{Main Results}
\begin{figure}[t]  
  \centering
  \includegraphics[width=\columnwidth]{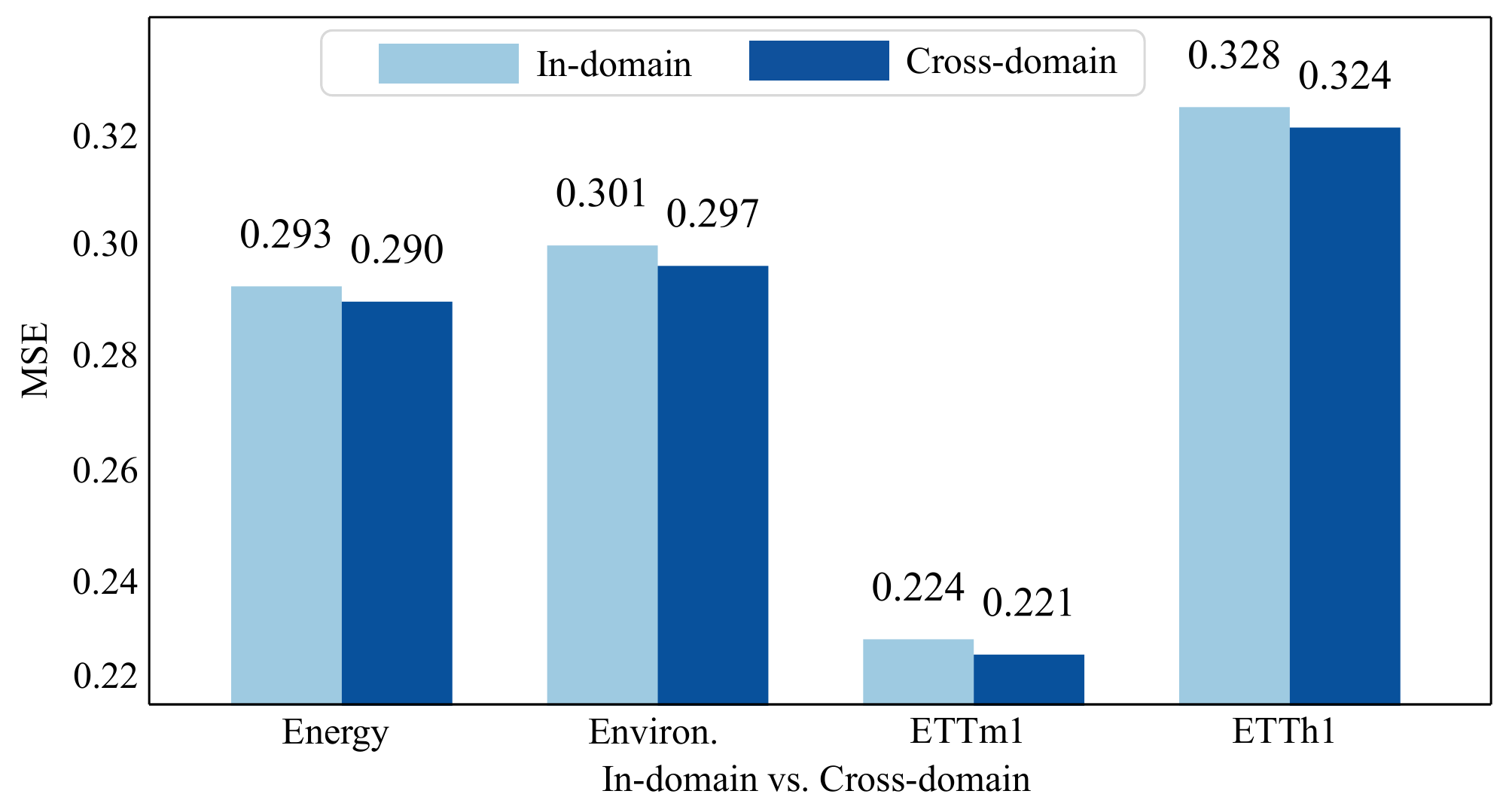} 
  
\caption{Performance comparison between in-domain and cross-domain training. The bar chart shows the MSE reduction  achieved by cross-domain training across four representative benchmarks.}
  \label{fig:crossdomain}
  
\end{figure}

Table~\ref{tab:mainresult} summarizes the main forecasting results evaluated across ten diverse, real-world benchmark datasets. Overall, CoGenCast achieves the  competitive performance compared to previous baselines, consistently delivering the most superior results across the vast majority of evaluation scenarios, which demonstrates strong  forecasting capability across diverse domains. In particular, our approach  outperforms representative LLM-based baselines (TokenCast, LLM4TS and Time-LLM), as well as strong generative-based and deep-based methods, suggesting that effective forecasting benefits from simultaneously capturing semantic understanding and continuous stochastic  dynamics. On average, we observe an approximate 10\% reduction in MSE compared to LLM-based approaches, a nearly 19\% reduction  compared to generative-based approaches and over 4\% improvement over strong deep-based baselines (e.g., ConvTimeNet), indicating that the gains stem from a coupled modeling of both structural dependencies and temporal dynamics rather than architectural scaling or parameter expansion alone.  In summary, CoGenCast delivers competitive performance with high consistency across varying prediction horizons. This advantage is enabled by structurally reconfiguring pre-trained decoder-only LLMs into a native forecasting encoder-decoder backbone via attention topology modification, and further integrating a flow-matching generative mechanism to model continuous stochastic dynamics conditioned on autoregressively generated representations. In addition to the in-domain evaluation in Table~\ref{tab:mainresult}, we further conduct a general cross-domain training setting to strengthen the generalization capability of our framework. As shown in Figure~\ref{fig:crossdomain}, cross-domain training leads to consistent performance improvements compared to training within a single domain, reducing MSE across multiple datasets (e.g., Energy, Environ., ETTm1, and ETTh1). These experimental results validate the effectiveness of our proposed CoGenCast.  Visualized results are provided in Appendix~\ref{sec:visulaiztion}, and the full results can be found in Appendix~\ref{sec:full result}.

\begin{table}[t]
\centering
\caption{Ablation study on the encoder-decoder architecture. We compare the forecasting performance of our full model with encoder-only and decoder-only variants.}
\label{tab:Encoder-Decoder Variant}

\setlength{\tabcolsep}{3pt} 
\renewcommand{\arraystretch}{1.2}
\small

\resizebox{\columnwidth}{!}{%
\begin{tabular}{l|cc|cc|cc|cc}
\toprule
\textsc{Methods} & \multicolumn{2}{c|}{\textsc{Energy}} & \multicolumn{2}{c|}{\textsc{Environ.}} & \multicolumn{2}{c|}{\textsc{Exchange}} & \multicolumn{2}{c}{\textsc{Wind}} \\
\textsc{Metrics} & MSE & MAE & MSE & MAE & MSE & MAE & MSE & MAE \\
\midrule
\textbf{Ours} & \textbf{0.293} & \textbf{0.395} & \textbf{0.301} & \textbf{0.381} & \textbf{0.031} & \textbf{0.118} & \textbf{0.532} & \textbf{0.425} \\
Encoder-only & 0.331 & 0.431 & 0.313 & 0.404 & 0.036 & 0.132 & 0.547 & 0.453 \\
Decoder-only & 0.953 & 0.742 & 0.602 & 0.574 & 0.188 & 0.255 & 2.010 & 0.833 \\
\bottomrule
\end{tabular}%
}
\end{table}

\subsection{Ablation Study}
\paragraph{Encoder-Decoder Variants.} To verify the effectiveness of our encoder-decoder architecture, we conduct experiments with encoder-decoder variants. For the encoder-only variant, we retain the  former encoder and add a linear  layer at the end of the model. For the decoder-only variant, we retain the latter decoder and the denoising decoder. As shown in Table~\ref{tab:Encoder-Decoder Variant}, the full encoder-decoder architecture achieves superior results compared to both variants on all datasets. Notably, the significant performance degradation in the decoder-only variant underscores the critical role of the encoder in providing semantic conditioning for the generative process. Meanwhile, the performance gap between the full architecture and the encoder-only variant verifies that integrating flow matching into the autoregressive optimization captures continuous dynamics more effectively than a linear  layer. Overall, these results confirm that combining bidirectional context modeling in the encoder with causal representation generation in the decoder is necessary, and the proposed encoder–decoder architecture is essential to achieve the superior forecasting performance.

\begin{table}[t]
\centering
\caption{Ablation study on the generative components. We evaluate the individual contributions of the autoregressive (AR)  and the flow-matching mechanism.}
\label{tab:ar_meanflow_ablation}
\setlength{\tabcolsep}{4pt}
\renewcommand{\arraystretch}{1.3}
\small
\resizebox{\columnwidth}{!}{%
\begin{tabular}{l|cc|cc|cc|cc}
\toprule
\textsc{Methods} & \multicolumn{2}{c|}{\textsc{Energy}} & \multicolumn{2}{c|}{\textsc{ETTh1}} & \multicolumn{2}{c|}{\textsc{ETTm1}} & \multicolumn{2}{c}{\textsc{Wind}} \\[2pt]
\textsc{Metrics} & MSE & MAE & MSE & MAE & MSE & MAE & MSE & MAE \\
\midrule
\textbf{Ours} & \textbf{0.293} & \textbf{0.395} & \textbf{0.328} & \textbf{0.367} & \textbf{0.224} & \textbf{0.285} & \textbf{0.532} & \textbf{0.425} \\
w/o AR & 0.339 & 0.427 & 0.366 & 0.402 & 0.257 & 0.322 & 0.673 & 0.501 \\
w/o Flow & 0.335 & 0.422 & 0.358 & 0.396 & 0.251 & 0.317 & 0.611 & 0.482 \\
w/o AR-Flow & 0.345 & 0.430 & 0.372 & 0.406 & 0.262 & 0.327 & 0.711 & 0.522 \\
\bottomrule
\end{tabular}%
}
\end{table}

\begin{figure}[!t]  
  \centering
  \includegraphics[width=\columnwidth]{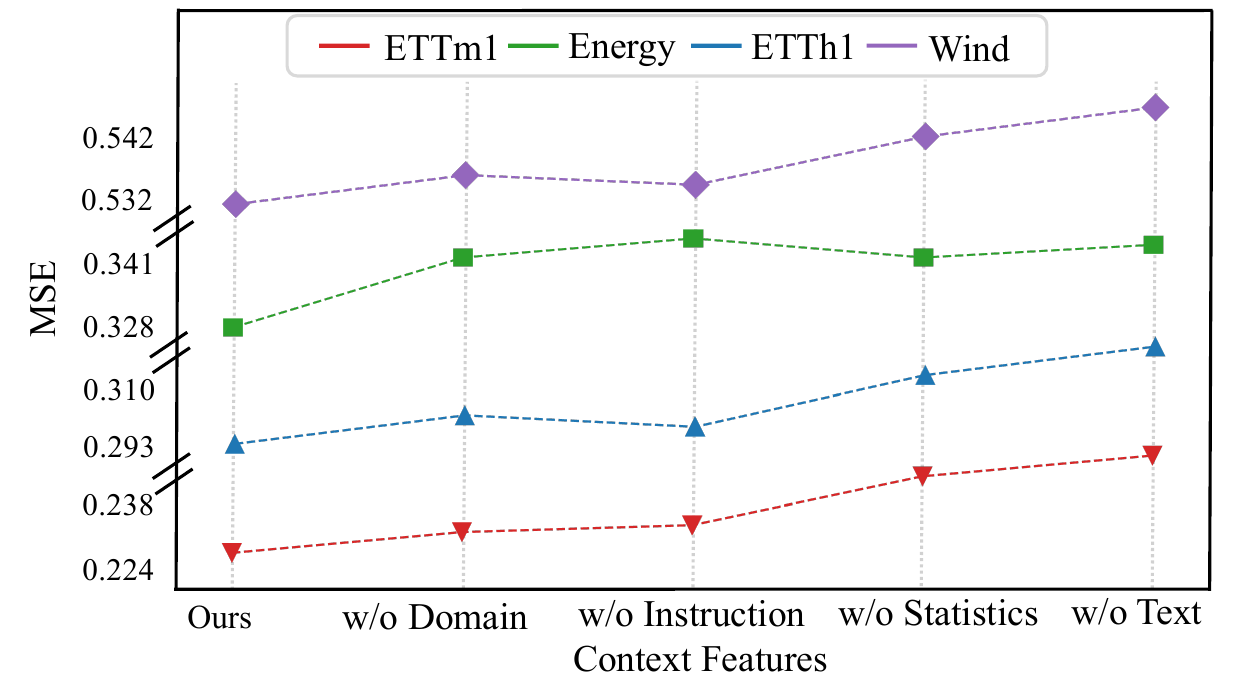} 

    \caption{Ablation study on context features. We evaluate the impact of removing domain knowledge, task instruction, statistics information, and  entire textual input on forecasting performance.}
  \label{fig:text_ablation}

\end{figure}

\paragraph{Ablation on AR-Flow.} Table~\ref{tab:ar_meanflow_ablation}  presents the ablation study results to evaluate the individual contributions of the autoregressive  mechanism and the flow-matching mechanism. We compare the full model  with three variants:  w/o AR,
w/o Flow and  w/o AR-Flow. The results  clearly show that removing either component leads to a significant degradation in forecasting accuracy. Specifically, the w/o AR variant exhibits a substantial increase in MSE, confirming that LLM-based semantic understanding is crucial for capturing long-term contextual dependencies. Simultaneously, the w/o Flow variant performs worse than the full model, validating that the flow-matching mechanism effectively models the continuous temporal dynamics in time series data. Ultimately, the full model achieves the best performance by successfully coupling these two methods, proving that structural semantic understanding and continuous stochastic modeling are both essential for forecasting performance.

\paragraph{Ablation on Context Features.} As shown in Figure~\ref{fig:text_ablation}, to  investigate the impact of specific context features on forecasting performance, we conduct a detailed ablation study. The considered variants include removing the entire textual input (w/o Text), as well as selectively excluding domain knowledge (w/o Domain), task instruction (w/o Instruction), and statistics information (w/o Statistics). The results indicate that the complete text-guided framework consistently yields the best performance. This confirms that the incorporation of  context features provides essential guidance, effectively enhancing  forecasting accuracy. Furthermore, it is particularly notable that removing the statistics information generally leads to a more pronounced error increase compared to other sub-components. Notably, all statistics information do not incorporate any future or target information, ensuring that no information leakage occurs.

\begin{table}[t]
\centering
\caption{Comparative analysis against generative baselines. The results demonstrate the superior performance of our method compared to vanilla flow-matching and diffusion models.}
\label{tab:generative_compare}
\setlength{\tabcolsep}{4pt}
\renewcommand{\arraystretch}{1.3}
\small
\resizebox{\columnwidth}{!}{%
\begin{tabular}{l|cc|cc|cc|cc}
\toprule
\textsc{Methods} & \multicolumn{2}{c|}{\textsc{Energy}} & \multicolumn{2}{c|}{\textsc{Environ.}} & \multicolumn{2}{c|}{\textsc{ETTh1}} & \multicolumn{2}{c}{\textsc{Exchange}} \\
\textsc{Metrics} & MSE & MAE & MSE & MAE & MSE & MAE & MSE & MAE \\
\midrule
\textbf{Ours} & \textbf{0.293} & \textbf{0.395} & \textbf{0.301} & \textbf{0.381} & \textbf{0.328} & \textbf{0.367} & \textbf{0.031} & \textbf{0.118} \\
Flow-matching  & 0.332 & 0.418 & 0.329 & 0.399 & 0.372 & 0.399 & 0.051 & 0.131 \\
Diffusion     & 0.325 & 0.413 & 0.322 & 0.395 & 0.363 & 0.394 & 0.046 & 0.127 \\
\bottomrule
\end{tabular}%
}
\end{table}

\begin{figure}[t]
    \centering
    \includegraphics[width=0.98\columnwidth]{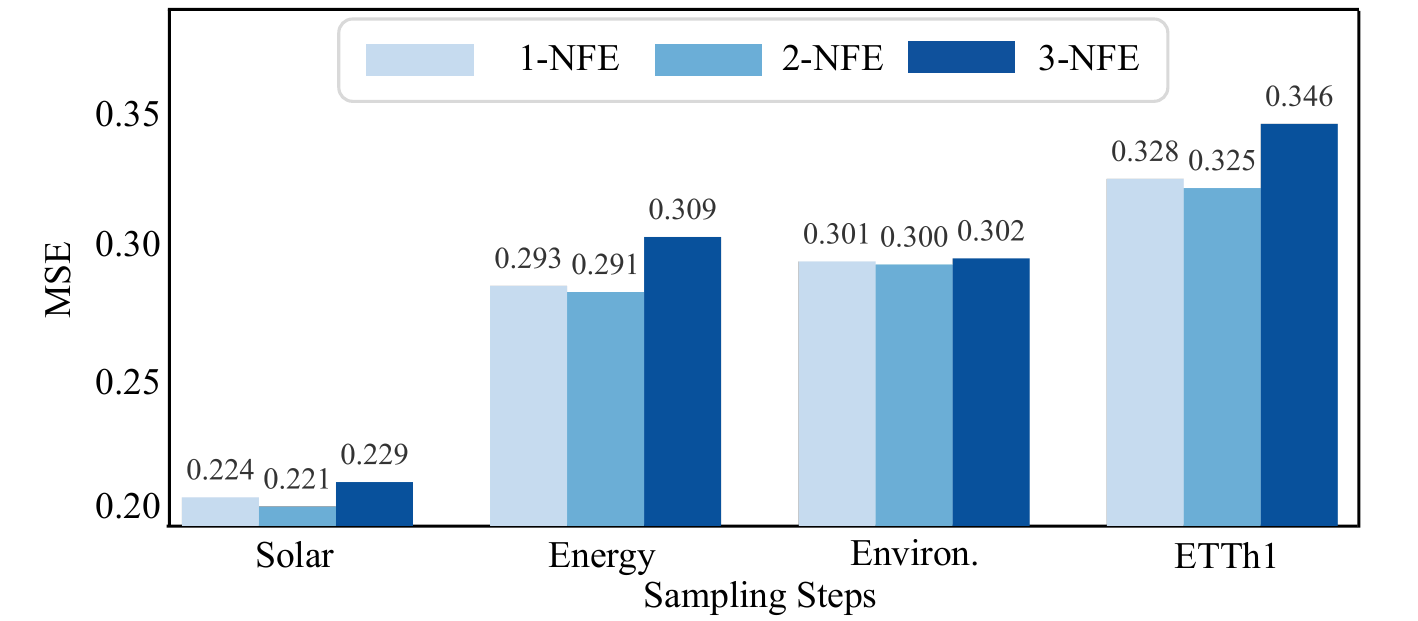}
\caption{Comparative analysis on the number of function evaluations (NFE). We compare the forecasting performance across 1, 2, and 3 sampling steps.}
    \label{fig:NFE_analysis}
\end{figure}

\subsection{Comparison with Generative Models}

\paragraph{Comparison with Generative Models.} To further evaluate the architectural advantages of our approach, we conduct a comparative study by replacing our generation module with  vanilla diffusion and flow-matching components within the identical framework. As demonstrated in Table~\ref{tab:generative_compare}, our method achieves superior forecasting performance across all benchmarks. This advantage is primarily attributed to the fundamental differences in modeling objectives. While diffusion models rely on complex iterative denoising to recover signals and standard flow-matching models target instantaneous velocity, our method explicitly models the average velocity connecting the prior to the ground truth. 

\begin{table}[t]
\centering
\caption{Efficiency analysis on the ETTh1 dataset. We evaluate the training and inference time (minutes) of our approach against the five representative baselines.}
\label{tab:efficiency}
\resizebox{\columnwidth}{!}{%
\setlength{\tabcolsep}{3pt} 
\renewcommand{\arraystretch}{1.3}
\begin{tabular}{l|c|c|c|c|c|c}
\toprule
\textsc{Time (min)} & Ours & \textsc{TokenCast} & \textsc{LLM4TS} & \textsc{FlowTS} & \textsc{CDPM} & \textsc{Autoformer} \\
\midrule
Training & 17.893 & 68.430 & 59.913 & 23.477 & 27.891 & \textbf{4.479} \\
Inference & 1.776 & 9.880 & 7.417 & 4.218 & 5.765 & \textbf{0.672} \\
\bottomrule
\end{tabular}%
}
\end{table}

\begin{table}[t]
\centering
\caption{Performance comparison across different backbone architectures. We evaluate the impact of backbone  by comparing the Qwen family (0.6B, 1.7B, 4B) with a vanilla Transformer.}
\label{tab:backbone}
\setlength{\tabcolsep}{4pt}
\renewcommand{\arraystretch}{1.3}
\small
\resizebox{\columnwidth}{!}{%
\begin{tabular}{l|cc|cc|cc|cc}
\toprule
\textsc{Backbone} & \multicolumn{2}{c|}{\textsc{Energy}} & \multicolumn{2}{c|}{\textsc{Environ.}} & \multicolumn{2}{c|}{\textsc{Exchange}} & \multicolumn{2}{c}{\textsc{Wind}} \\[2pt]
\textsc{Metrics} & MSE & MAE & MSE & MAE & MSE & MAE & MSE & MAE \\
\midrule
Transformer & 0.301 & 0.399 & 0.311 & 0.388 & 0.037 & 0.123 & 0.587 & 0.461 \\
Qwen3-0.6B  & 0.293 & 0.395 & 0.301 & 0.381 & 0.031 & 0.118 & 0.532 & 0.425 \\
Qwen3-1.7B  & 0.295 & 0.396 & 0.300 & 0.380 & 0.031 & 0.117 & 0.530 & 0.424 \\
Qwen3-4B    & \textbf{0.290} & \textbf{0.392} & \textbf{0.297} & \textbf{0.376} & \textbf{0.029} & \textbf{0.115} & \textbf{0.525} & \textbf{0.417} \\
\bottomrule
\end{tabular}%
}
\end{table}

\begin{figure}[t]
    \centering
    \includegraphics[width=0.98\columnwidth]{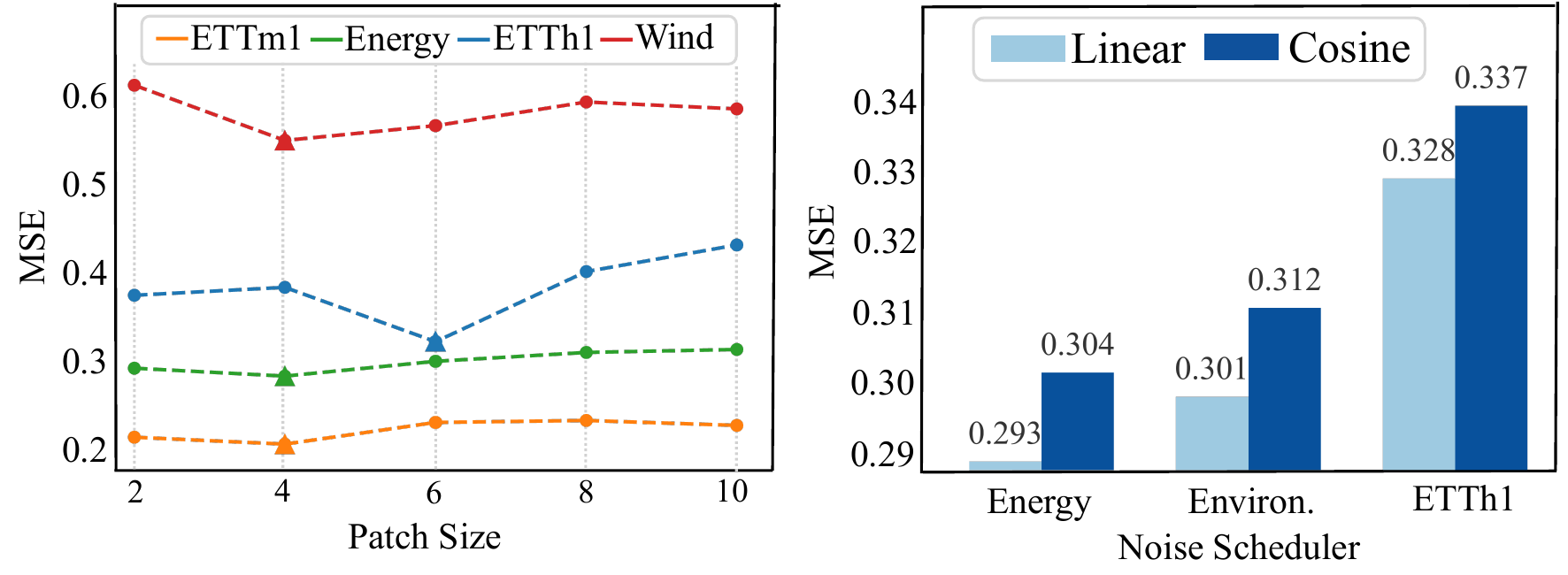}
    \caption{Hyperparameter sensitivity analysis. (Left) Impact of varying patch sizes on MSE across different datasets. (Right) Performance comparison between linear and cosine noise schedules.}
    \label{fig:Hyper_analysis}
\end{figure}

\begin{table}[t]
\centering
\caption{Performance comparison across different model backbones. We evaluate the predictive accuracy of the default Qwen3-0.6B against the Llama3.2-1B variant.}
\label{tab:llama_backbone}
\resizebox{\columnwidth}{!}{%
\setlength{\tabcolsep}{6pt}
\renewcommand{\arraystretch}{1.3}
\small
\begin{tabular}{l|cc|cc|cc|cc}
\toprule
\textsc{Backbone} & \multicolumn{2}{c|}{\textsc{ETTh2}} & \multicolumn{2}{c|}{\textsc{ETTm1}} & \multicolumn{2}{c|}{\textsc{ETTm2}} & \multicolumn{2}{c}{\textsc{Exchange}} \\
\textsc{Metrics} & MSE & MAE & MSE & MAE & MSE & MAE & MSE & MAE \\
\midrule
Ours(Qwen3-0.6B) & 0.159 & \textbf{0.246} & 0.224 & 0.285 & 0.109 & 0.200 & 0.031 & 0.118 \\
Llama3.2-1B & \textbf{0.158 }& 0.247 & \textbf{0.221} & \textbf{0.283} & \textbf{0.108} & \textbf{0.198} & \textbf{0.030} & \textbf{0.116}\\
\bottomrule
\end{tabular}%
}
\end{table}

\paragraph{Efficiency-Performance Trade-off.} As shown in Figure~\ref{fig:NFE_analysis}, we investigate the impact of the number of function evaluations (NFE) on forecasting performance. Remarkably, one-step (1-NFE) generation achieves performance close to the best 2-NFE setting, whereas extending the inference to 3-NFE results in slight marginal degradation. This confirms that our learned temporal trajectories are nearly linear, making a single Euler step sufficient for high-precision generation. Additional steps become redundant and prone to numerical error accumulation. Therefore, we adopt 1-NFE to achieve the best trade-off, ensuring real-time inference speed without compromising performance.

\paragraph{One-step Efficiency Analysis.} 
To further evaluate the practical utility of CoGenCast, we conduct a quantitative analysis of computational efficiency on the ETTh1 dataset. 
As presented in Table~\ref{tab:efficiency}, CoGenCast demonstrates a clear speed advantage over the representative LLM-based and generative-based baselines. 
Our framework exhibits  lower inference latency compared to leading LLM-based methods such as TokenCast and LLM4TS. 
This efficiency is primarily attributed to our one-step generative modeling mechanism, which allows the temporal evolution to approximate the entire transport process in a single function evaluation. 
By linearizing the generative trajectory, CoGenCast bypasses the iterative denoising process inherent in vanilla diffusion-based models like CDPM. 
While traditional deep-based architectures remain faster due to their more compact parameter scales, CoGenCast achieves a  performance-efficiency trade-off, delivering the strong forecasting accuracy with a highly competitive operational overhead.

\subsection{Hyperparameter Sensitivity}
\paragraph{Patch Size.} As visualized in Figure~\ref{fig:Hyper_analysis}, we investigate the impact of patch size on forecasting performance. The results demonstrate that different datasets exhibit varying levels of sensitivity to this hyperparameter. Notably, neither an overly small nor an excessively large patch size is desirable. This is attributed to the fact that the former fails to capture sufficient local semantic context, while the latter risks smoothing out fine-grained temporal dynamics. Thus, selecting a suitable patch size is essential to the forecasting performance.

\paragraph{Noise Scheduler.} We investigate the impact of different noise scheduler strategies on forecasting performance, comparing linear and cosine schedulers. As shown in Figure~\ref{fig:Hyper_analysis}, the linear scheduler yields superior results across most datasets. This is attributed to the linear scheduler offering uniform discretization that perfectly aligns with our average velocity modeling, which implies a straight constant-speed trajectory. Crucially, this alignment is significantly more suitable for our one-step generation modeling, as it ensures precise inference in a single step without the unnecessary temporal curvature introduced by the cosine scheduler.


\subsection{Analysis Experiment}
\paragraph{LLM Backbone.} To verify the effectiveness of the backbone architecture, we conduct a comparative analysis between a vanilla  Transformer and the Qwen LLM family across different parameter scales, as shown in Table~\ref{tab:backbone}. The results demonstrate that LLM-based backbones significantly outperform the vanilla  Transformer backbone. This is attributed to the LLMs' stronger fitting capability and the integration of multi-modal context features, which provide deeper semantic understanding to explicitly condition the generation process. Notably, comparing within the Qwen family, the Qwen3-4B model achieves the best performance. However, the performance gap between the 0.6B and 4B models is marginal. Considering the comprehensive trade-off between computational resources, training time, and forecasting performance, we selected Qwen3-0.6B as our default backbone, as it delivers competitive performance with substantially higher efficiency.  Furthermore, as shown in Table~\ref{tab:llama_backbone}, we additionally evaluate our framework by comparing the  backbone with Llama3.2-1B across four representative datasets. The experimental results demonstrate that CoGenCast still achieves state-of-the-art performance, with the Llama-based variant showing slight improvements in several benchmarks. This consistent performance across different LLM families  suggests that our architectural design is not tied to a specific backbone, demonstrating its generality.



\begin{table}[t]
\centering
\caption{Quantitative comparison of generative uncertainty modeling performance. We evaluate the predictive uncertainty by comparing our approach against NsDiff and TMDM.}
\label{tab:uncertainty_quant}
\resizebox{\columnwidth}{!}{%
\setlength{\tabcolsep}{4pt}
\renewcommand{\arraystretch}{1.3}
\small
\begin{tabular}{l|cc|cc|cc|cc}
\toprule
\textsc{Methods} & \multicolumn{2}{c|}{\textsc{ETTh2}} & \multicolumn{2}{c|}{\textsc{ETTm1}} & \multicolumn{2}{c|}{\textsc{ETTm2}} & \multicolumn{2}{c}{\textsc{Exchange}} \\
\textsc{Metrics} & CRPS & QICE & CRPS & QICE & CRPS & QICE & CRPS & QICE \\
\midrule
\textbf{Ours} & \textbf{0.122$_{\pm 0.003}$} & \textbf{0.021$_{\pm 0.001}$} & \textbf{0.169$_{\pm 0.004}$} & \textbf{0.023$_{\pm 0.002}$} & \textbf{0.088$_{\pm 0.002}$} & \textbf{0.015$_{\pm 0.001}$} & \textbf{0.024$_{\pm 0.001}$} & \textbf{0.011$_{\pm 0.001}$} \\
NsDiff & 0.145$_{\pm 0.006}$ & 0.032$_{\pm 0.003}$ & 0.198$_{\pm 0.007}$ & 0.035$_{\pm 0.004}$ & 0.112$_{\pm 0.005}$ & 0.024$_{\pm 0.002}$ & 0.038$_{\pm 0.003}$ & 0.019$_{\pm 0.002}$ \\
TMDM   & 0.138$_{\pm 0.005}$ & 0.028$_{\pm 0.002}$ & 0.185$_{\pm 0.006}$ & 0.031$_{\pm 0.003}$ & 0.105$_{\pm 0.004}$ & 0.022$_{\pm 0.002}$ & 0.033$_{\pm 0.002}$ & 0.016$_{\pm 0.001}$ \\
\bottomrule
\end{tabular}%
}
\end{table}

\begin{figure}[t]
    \centering
    \includegraphics[width=0.99\columnwidth]{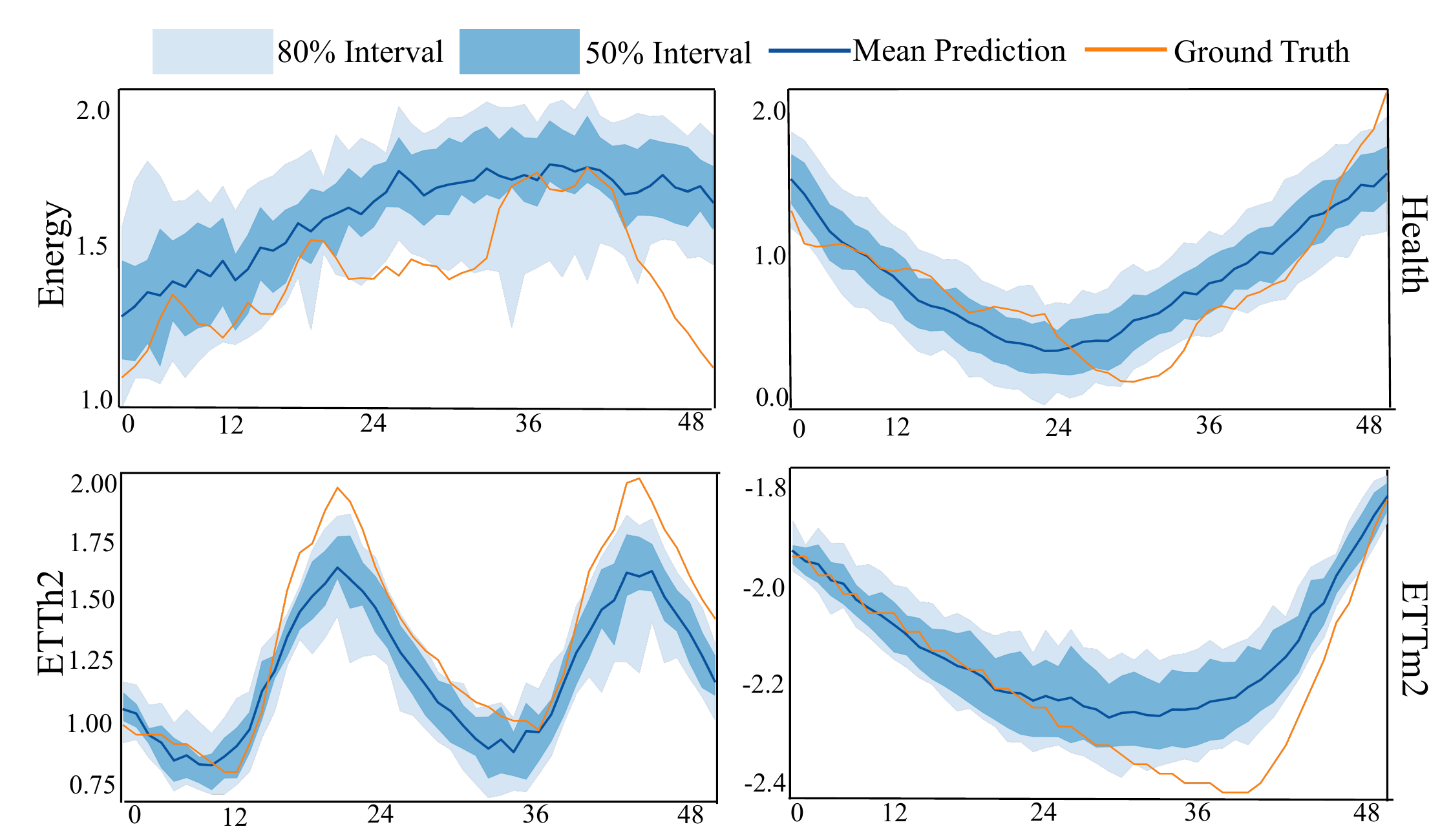}
    \caption{Forecasting with uncertainty on four datasets. The plots compare the ground truth trajectories with the model’s mean predictions, along with the 50\% and 80\% predictive intervals.}
    \label{fig:Uncertainty}
\end{figure}

\begin{table}[t]
\centering
\caption{Scalability analysis regarding the look-back window length $L$. We evaluate the predictive accuracy of our approach under different input scales, specifically $L \in \{96, 192, 336\}$.}
\label{tab:scalability}
\resizebox{\columnwidth}{!}{%
\setlength{\tabcolsep}{6pt}
\renewcommand{\arraystretch}{1.3}
\small
\begin{tabular}{l|cc|cc|cc|cc}
\toprule
\textsc{Look-back} & \multicolumn{2}{c|}{\textsc{Energy}} & \multicolumn{2}{c|}{\textsc{Environ.}} & \multicolumn{2}{c|}{\textsc{ETTh1}} & \multicolumn{2}{c}{\textsc{ETTm1}} \\
\textsc{metrics} & MSE & MAE & MSE & MAE & MSE & MAE & MSE & MAE \\
\midrule
96  & 0.293 & 0.395 & 0.301 & 0.381 & 0.328 & 0.367 & 0.224 & 0.285 \\
192 & 0.290 & 0.393 & 0.298 & 0.379 & 0.326 & 0.366 & 0.223 & 0.284 \\
336 & \textbf{0.288} & \textbf{0.390} & \textbf{0.280} & \textbf{0.375} & \textbf{0.321} & \textbf{0.362} & \textbf{0.219} & \textbf{0.281} \\
\bottomrule
\end{tabular}%
}
\end{table}

\begin{table}[t]
\centering
\caption{Long-term forecasting performance comparison across extended horizons. We evaluate the average results of our approach against three representative baselines.}
\label{tab:long_term_horizons}
\resizebox{\columnwidth}{!}{%
\setlength{\tabcolsep}{6pt}
\renewcommand{\arraystretch}{1.3}
\small
\begin{tabular}{l|cc|cc|cc|cc}
\toprule
\textsc{Methods} & \multicolumn{2}{c|}{\textsc{Energy}} & \multicolumn{2}{c|}{\textsc{Environ.}} & \multicolumn{2}{c|}{\textsc{Exchange}} & \multicolumn{2}{c}{\textsc{Health}} \\
\textsc{Metrics} & MSE & MAE & MSE & MAE & MSE & MAE & MSE & MAE \\
\midrule
\textbf{Ours} & \textbf{0.601} & \textbf{0.606} & \textbf{0.336} & \textbf{0.408} & \textbf{0.398} & \textbf{0.442} & \textbf{1.890} & \textbf{0.953} \\
TokenCast & 0.612 & 0.623 & 0.341 & 0.412 & 0.455 & 0.459 & 1.991 & 0.969 \\
FlowTS & 0.646 & 0.641 & 0.369 & 0.431 & 0.471 & 0.487 & 2.079 & 0.981 \\
TimeDART & 0.610 & 0.621 & 0.340 & 0.410 & 0.444 & 0.468 & 1.901 & 0.961 \\
\bottomrule
\end{tabular}%
}
\end{table}

\paragraph{Generative Uncertainty.}  To validate the uncertainty modeling capabilities of our framework, we visualize the different predictive distributions in Figure~\ref{fig:Uncertainty}. The results demonstrate that our method generates forecasting distributions that closely align with the ground truth, with the 50\% and 80\% confidence intervals effectively encompassing the inherent data variability. Such reliability suggests that our approach excels at capturing the stochastic nature of time series, offering significant practical potential for  uncertainty-aware forecasting. Furthermore, to provide a more rigorous quantitative evaluation of generative uncertainty, we compare CoGenCast with leading generative baselines, NsDiff~\cite{ye2025non} and TMDM~\cite{li2024transformer}, using CRPS~\cite{ye2025non} and QICE~\cite{ye2025non} metrics. For a fair comparison, each method is sampled 20 times for every forecasting instance, and the uncertainty metrics are computed from the resulting predictive distribution. As shown in Table~\ref{tab:uncertainty_quant}, CoGenCast consistently achieves the competitive performance across evaluated datasets, indicating its superior ability to capture the underlying probability distribution of time series. Moreover, our method maintains competitive QICE scores, demonstrating better calibration in uncertainty estimation. These results show the advantages of our coupled  design in generating  probabilistic forecasts compared to traditional diffusion-based generative models.

\paragraph{Scalability.}
Our framework is designed to support variable-length look-back windows and extended forecasting horizons without structural modifications. To validate this capability and evaluate the model's ability to leverage long-term historical context, we first investigated the impact of progressively extending the look-back window size $L$ from 96 to 192 and 336. As shown in our evaluation, the forecasting performance of CoGenCast improves consistently as the historical context expands. This suggests that our reconfigured LLM-based encoder effectively distills deep semantic patterns from increasingly dense temporal sequences, successfully mapping complex long-range dependencies into the latent representation space.
And we evaluate the scalability of CoGenCast under challenging long-term forecasting scenarios with extended horizons $H \in \{192, 336, 720\}$. As demonstrated in Table~\ref{tab:long_term_horizons}, our method maintains a decisive performance advantage over competitive baselines, including TokenCast, FlowTS, and TimeDART. While conventional iterative generative models often suffer from cumulative error propagation over elongated prediction intervals, CoGenCast’s coupled autoregressive-flow mechanism ensures stable and precise temporal trajectories by modeling the global average velocity rather than instantaneous local shifts. This  scalability confirms that CoGenCast is suited for complex applications.

\section{Conclusion}

In this paper, we proposed CoGenCast, a hybrid generative framework that coupled pre-trained LLMs with a flow-matching mechanism for effective time series forecasting. This bridged the critical gap between semantic understanding and continuous stochastic modeling. By reconfiguring pre-trained decoder-only LLMs into a native forecasting encoder-decoder backbone, we leveraged comprehensive contextual understanding to condition the flow-matching generation process. Our approach linearized the generative dynamics, establishing a direct and stable temporal trajectory that could be solved efficiently with a single function evaluation. Extensive experiments on ten benchmark datasets showed that CoGenCast achieves  competitive performance compared to previous  baselines. Finally, we hope that this work can offer a valuable and inspiring perspective for future time series forecasting methods.

\section*{Acknowledgments}
This work was supported by grants from the National Natural Science Foundation of China (No. 62502486,U25B2072), the Key Technologies R \& D Program of Anhui Province (No. 202423k09020039), the Fundamental Research Funds for the Central Universities of China (No. WK2150110032), USTC Research Funds of the DoubleFirst-Class Initiative (No. YD2150002501).

\section*{Impact Statement}
This paper introduces a hybrid generative framework that couples  pre-trained  LLMs with  flow-matching mechanism for effective time series forecasting. Our work offers valuable insights for future research in the synergy between LLMs and continuous probabilistic generation. Experimental results demonstrate the effectiveness of our method and its potential real-world applicability. We have ensured that all datasets used in the experiments are publicly available, and we have carefully considered the ethical implications of our work. In practical deployment, CoGenCast should be used with appropriate domain expert validation, uncertainty assessment, and safety monitoring. 
Its predictions should serve as decision-support information rather than the sole basis for high-stakes decisions.

\bibliography{icml}
\bibliographystyle{icml2026}

\newpage
\appendix
\onecolumn
\section{Dataset Description}
\label{sec:dataset description}
To evaluate the effectiveness of our proposed framework, we conduct extensive experiments on ten time series forecasting datasets. These datasets cover diverse domains, including energy, environment,  finance, health and electricity. For detailed description of the datasets, please refer to Table~\ref{tab:dataset_full}.

\begin{table}[h]
\centering
\caption{Full dataset descriptions. Samples are organized in (Train/Validation/Test).}
\label{tab:dataset_full}

\setlength{\tabcolsep}{6pt}
\renewcommand{\arraystretch}{1.3} 
\small

\resizebox{0.95\textwidth}{!}{%
\begin{tabular}{lccccccc}
\toprule
\textsc{Datasets} & \textsc{Look-back} & \textsc{Predicted} & \textsc{Variables} & \textsc{Samples} & \textsc{Domain} & \textsc{Frequency} \\
\midrule
Energy      & 96 & $\{12,24,36,48\}$ & 1   & 928/138/284       & Energy       & 1 Week \\
ETTh1       & 96 & $\{12,24,36,48\}$ & 7   & 8293/2869/2869    & Energy       & 1 Hour \\
ETTh2       & 96 & $\{12,24,36,48\}$ & 7   & 8293/2869/2869    & Energy       & 1 Hour \\
ETTm1       & 96 & $\{12,24,36,48\}$ & 7   & 34417/11473/11473 & Energy       & 15 Mins \\
ETTm2       & 96 & $\{12,24,36,48\}$ & 7   & 34417/11473/11473 & Energy       & 15 Mins \\
Environment & 96 & $\{12,24,36,48\}$ & 1   & 10566/1515/3038   & Environment  & 1 Day \\
Exchange    & 96 & $\{12,24,36,48\}$ & 8   & 4928/713/1470     & Finance      & 1 Day \\
Health      & 96 & $\{12,24,36,48\}$ & 1   & 829/93/230        & Health       & 1 Week \\
Wind        & 96 & $\{12,24,36,48\}$ & 7   & 33688/4821/9687   & Electricity  & 15 Mins \\
Solar       & 96 & $\{12,24,36,48\}$ & 137 & 36445/5245/10501  & Electricity  & 10 Mins \\
\bottomrule
\end{tabular}%
}
\end{table}

\paragraph{Energy~\cite{liu2024time}} This dataset comprises weekly U.S. gasoline price time series spanning from April 1993 to the present. The data is officially released by the U.S. Energy Information Administration (EIA)
\paragraph{ETT(4 subsets)~\cite{zhou2021informer}} This dataset comprises time series data of oil temperature and power load collected
from electricity transformers spanning July 2016 to July 2018. It is divided into four subsets, each with different recording
intervals: ETTh1 and ETTh2 have hourly recordings, while ETTm1 and ETTm2 are recorded every 15 minutes.

\paragraph{Environment~\cite{liu2024time}} This dataset comprises daily air quality time series collected from the U.S. EPA spanning the full available period, recorded at a daily frequency.

\paragraph{Exchange~\cite{wu2021autoformer}}This dataset collects the daily exchange rates of eight countries—Australia, the UK,
Canada, Switzerland, China, Japan, New Zealand, and Singapore—from 1990 to 2016.
\paragraph{Health~\cite{liu2024time}} This dataset comprises weekly U.S. Influenza-Like Illness (ILI) statistics spanning from 1954 to the present, recorded at a weekly frequency. The accompanying data is collected from the CDC Weekly U.S. Influenza Surveillance Report, released weekly.

\paragraph{Wind~\cite{li2022generative}}This dataset comprises wind power measurements sampled every 15 minutes from 2020 to 2021.
\paragraph{Solar~\cite{lai2018modeling}}This dataset records the solar power production of 137 PV plants in 2006, which are sampled every 10 minutes. 

\section{Implementation Details}
\label{sec:implementaion details}
\subsection{Compared Baselines}
\label{app:baselines}
To more comprehensively evaluate the capabilities of our proposed method, we compare our method with thirteen representative baselines in our experiments.  Specifically, we include LLM-based approaches (TokenCast, LLM4TS and Time-LLM), generative-based models (FlowTS, CDPM, and TimeGrad), and deep-based  methods (ConvTimeNet, TimeDART, iTransformer, TimesNet, PatchTST, DLinear, and Autoformer).

\paragraph{TokenCast~\cite{tao2025values}}TokenCast\footnote{\url{https://github.com/ustc-time-series/TokenCast}} introduces a novel framework that reformulates time series forecasting as a next-token prediction task, allowing language models to predict future values auto-regressively.
\paragraph{LLM4TS~\cite{chang2025llm4ts}}LLM4TS\footnote{\url{https://github.com/blacksnail789521/LLM4TS}} is a framework that adapts pre-trained LLMs for time series forecasting via two-stage fine-tuning (time-series alignment and downstream forecasting) and a two-level multi-scale temporal aggregation mechanism, enabling LLMs to better capture time series characteristics and multi-resolution patterns, especially in data-scarce settings.
\paragraph{Time-LLM~\cite{jin2024time}}Time-LLM\footnote{\url{https://github.com/KimMeen/Time-LLM}} is a reprogramming framework that aligns time-series inputs with text prototypes and feeds them into a frozen LLM, enhanced by a Prompt-as-Prefix mechanism to guide temporal reasoning for forecasting.

\paragraph{FlowTS~\cite{hu2024flowts}}FlowTS\footnote{\url{https://github.com/UNITES-Lab/FlowTS}} is an ODE-based rectified-flow model that learns straight-line (geodesic) transport paths in probability space for efficient exact trajectory simulation, further enhanced with adaptive sampling and time-series decomposition plus global-context/positional modules to improve generation quality.

\paragraph{CDPM~\cite{zhang2025conditional}}CDPM\footnote{\url{https://github.com/zjt-gpu/CDPM}} is a hybrid framework that decomposes time series into trend and seasonal components and trains a diffusion-based denoiser for stochastic seasonal fluctuations alongside an enhanced linear model for smooth trends, jointly optimized end-to-end to capture both complex dynamics and long-term structure.

\paragraph{TimeGrad~\cite{rasul2021autoregressive}}TimeGrad\footnote{\url{https://github.com/zalandoresearch/pytorch-ts}} employs an autoregressive diffusion framework to learn the continuous data distribution for probabilistic time series forecasting.

\paragraph{ConvTimeNet~\cite{cheng2025convtimenet}}ConvTimeNet\footnote{\url{https://github.com/ustc-time-series/ConvTimeNet}} utilizes a hierarchical pure convolutional architecture with a deformable patch layer to adaptively learn local patterns and multi-scale temporal dependencies.

\paragraph{TimeDART~\cite{wang2025timedart}}TimeDART\footnote{\url{https://github.com/Melmaphother/TimeDART}} is a self-supervised pre-training framework that combines causal transformer-based autoregressive modeling with a denoising diffusion process to jointly learn global trend evolution and fine-grained local patterns for transferable time-series representations.

\paragraph{iTransformer~\cite{liu2024itransformer}}iTransformer\footnote{\url{https://github.com/thuml/iTransformer}} inverts the traditional Transformer architecture to embed entire time series as individual tokens, effectively capturing multivariate correlations.

\paragraph{TimesNet~\cite{wu2022timesnet}}TimesNet\footnote{\url{https://github.com/thuml/TimesNet}} addresses general time series analysis by reshaping 1D sequences into 2D tensors, extending the capability of 2D CNNs to temporal representation learning.

\paragraph{PatchTST~\cite{nie2022time}}PatchTST\footnote{\url{https://github.com/yuqinie98/PatchTST}} is a patch-based, channel-independent Transformer that treats subseries patches as tokens and shares model weights across variables, greatly reducing attention cost while preserving local semantics to improve long-horizon forecasting and self-supervised representation learning.

\paragraph{DLinear~\cite{zeng2023transformers}}DLinear\footnote{\url{https://github.com/cure-lab/LTSF-Linear}} decomposes time series into trend and seasonal components, applying simple linear layers to each for efficient and robust forecasting.

\paragraph{Autoformer~\cite{wu2021autoformer}}Autoformer\footnote{\url{https://github.com/thuml/Autoformer}} is a decomposition-based forecasting architecture that embeds progressive trend/seasonal decomposition into the model and replaces self-attention with an Auto-Correlation mechanism to capture periodic dependencies and aggregate representations at the sub-series level for efficient long-horizon prediction.

\subsection{Implementation Details for Experiment}
\label{app:implementation}
The look-back window length was set to $L = 96$, and the prediction window lengths were set to $H \in \{12, 24, 36, 48\}$ for all datasets. 
The LLM encoder and decoder were initialized from the same Qwen3-0.6B checkpoint but were not weight-shared. 
For the continuous flow-matching mechanism, we adopted a linear noise scheduler. 
All model parameters were updated during training under a full fine-tuning setting. 
To ensure consistency, a batch size of 4 and a channel-independent configuration were applied across all datasets. 
We employed the Adam optimizer with an initial learning rate selected from $\{5 \times 10^{-4}, 10^{-4}, 5 \times 10^{-5}\}$. 
The training process was conducted for 10 epochs.

We adopted Mean Squared Error (MSE) and Mean Absolute Error (MAE) as our primary evaluation metrics. 
All baseline methods were evaluated using their respective official implementations to ensure a fair comparison. 
The entire framework was implemented in PyTorch and executed on a single NVIDIA A100 GPU.

\section{Flow Matching Details}

We provide additional details on the interval-conditioned average velocity objective used in our flow-matching module. 
Unlike standard flow-matching methods that estimate the instantaneous velocity field at a single time point, our formulation estimates the average velocity over a finite interval. 
Given a clean future patch $y$, Gaussian noise $\epsilon \sim \mathcal{N}(0,I)$, and an interpolation time $t \in [0,1]$, we construct the linear probability path:
\begin{equation}
\hat{y}_t = (1-t)\epsilon + t y,
\end{equation}
where the corresponding base displacement is:
\begin{equation}
v = y - \epsilon.
\end{equation}
For a target time $r \sim \mathcal{U}[t, 1]$, the denoising decoder predicts an interval-conditioned velocity:
\begin{equation}
u_\theta(\hat{{z}}^{{in}}_j,t,r,{z}^{{dec\_out}}_{1:j}),
\end{equation}
where $\hat{{z}}^{{in}}_j$ denotes the embedding of the noisy patch and ${z}^{{dec\_out}}_{1:j}$ denotes the autoregressive condition produced by the LLM decoder.
The average velocity over the interval $[t,r]$ can be related to the local velocity field through a first-order correction. 
Therefore, instead of directly regressing the predicted velocity to the base displacement $v$, we use a JVP-corrected target:
\begin{equation}
\tilde{v} = v - (r-t)\frac{\partial u_\theta}{\partial t},
\end{equation}
where  $\frac{\partial u_\theta}{\partial t}$ denotes the derivative of the predicted velocity with respect to the scalar time input. 
In implementation, this derivative is computed using automatic differentiation as a Jacobian-vector product (JVP) with respect to the time variable. 
The corrected target is treated as a stop-gradient target when computing the regression loss, which avoids explicitly back-propagating through the target construction and prevents unnecessary higher-order optimization overhead. 
The training objective is written as:
\begin{equation}
\mathcal{L}_{\mathrm{flow}}
=
\mathbb{E}_{t,r,\epsilon,y}
\left[
\frac{1}{N}
\sum_{j=1}^{N}
\left\|
u^{out}_j - \mathrm{sg}(\tilde{v}_j)
\right\|_2^2
\right],
\end{equation}
where $\mathrm{sg}(\cdot)$ denotes the stop-gradient operation. 
This objective encourages the denoising decoder to estimate the interval-level average velocity, reducing the error caused by relying only on local instantaneous velocity estimation and making the generation process more suitable for few-step or one-step sampling.

During inference, the continuous transport from noise to the clean prediction can be written in integral form:
\begin{equation}
y^{out}_j = y^{(0)}_j + \int_0^1 u_\theta(z_\tau,\tau,z^{dec\_out}_{1:j}) d\tau.
\end{equation}

Since the denoising decoder is trained to estimate the average velocity over a finite interval, we approximate the above integral with a single interval-conditioned prediction by setting $t=0$ and $r=1$:
\begin{equation}
y^{out}_j
=
y^{(0)}_j
+
u_\theta(z^{(0)}_j,0,1,z^{dec\_out}_{1:j}).
\end{equation}
This can be interpreted as a single Euler step over the interval $[0,1]$, or equivalently as directly estimating the integral displacement using the learned average velocity. 
Therefore, CoGenCast requires only one function evaluation within each autoregressive patch generation step, avoiding iterative denoising or multi-step ODE solving while preserving causal dependence across future patches.

\section{Full Results}
\label{sec:full result}
Due to space limitations, the complete tables for the full results  are provided in Appendix~\ref{sec: tables}. Experiments in the in-domain setting are shown in Table~\ref{tab:full_main_result}, while experiments in the cross-domain setting are presented in Table~\ref{tab:full_cross_domain_result}. 

Specifically, to evaluate generalization of our CoGenCast across domains, we conduct cross-domain forecasting experiments. 
In the full cross-domain setting, a unified model is trained on the combined training splits of multiple datasets and evaluated on each target domain separately.  Cross-domain training  improves performance over in-domain training, demonstrating the benefit of shared temporal patterns and contextual knowledge across datasets.  In the few-shot cross-domain setting, the model is first pre-trained on source datasets (except the target dataset) and then fine-tuned on a small fraction (5\%, 10\%, 20\%) of the target dataset (Table~\ref{tab:cross_domain_fewshot}). 
Even with very limited target data, the model achieves substantial accuracy gains, showing that cross-domain pretraining provides  data-efficient adaptation.

\section{Ablation Study Full Results}
\label{sec: full abltion}
\subsection{Encoder-Decoder Variants} As shown  in Table~\ref{tab:main_results_compact}, to validate the architectural design of our framework, we performed a comparative analysis between the complete encoder-decoder model and its two primary variants: Encoder-only and Decoder-only architectures, implemented with varying model sizes (0.6B and 1.7B). As presented in the ablation results, the full architecture  yields superior forecasting accuracy across most benchmark datasets. The Decoder-only variants exhibit significant performance degradation, particularly on datasets with complex temporal patterns such as Health and Wind. This substantial drop in performance confirms that the bidirectional encoder is essential for capturing deep semantic representations and historical context, which are critical for effective time series forecasting. Crucially, our analysis reveals that the performance advantage of our proposed architecture is not merely a result of increased parameter count. Experiments prove that our full model consistently outperforms the Encoder-only variant, even when the latter is scaled to a significantly larger 1.7B backbone. This demonstrates that the structural synergy between bidirectional contextual understanding and causal generative refinement is the primary driver of our model's effectiveness, offering consistent improvements.

\subsection{Ablation on AR-Flow} As shown  in Table~\ref{tab:ablation_study}, to investigate the individual contributions of the two core generative components, we conducted a granular ablation study across all ten benchmarks by comparing the full framework with three restricted variants: w/o AR, which removes the bidirectional encoding and causal generation backbone; w/o Flow, which replaces the continuous flow-matching process with a deterministic output; and w/o AR-Flow, representing a baseline without either advancement. As shown in the ablation results, the full model consistently achieves superior precision, while the removal of either module leads to a noticeable performance decay across varying horizons. Specifically, the w/o AR variant exhibits a marked increase in both MSE and MAE, confirming that the LLM-based semantic reasoning and structural prior are fundamental for capturing complex, long-range temporal dependencies. Simultaneously, the w/o Flow variant yields inferior results compared to the full model, validating that the continuous flow-matching module is essential for effectively modeling the stochasticity and fine-grained dynamics  in continuous-valued time series. The significant degradation observed in the w/o AR-Flow variant further underscores that the synergy between discrete semantic coupling and linearized continuous generation is the primary driver of our model's superior forecasting stability.

\subsection{Ablation on Context Features} 
In this work, multimodal forecasting refers to a practical numerical-textual forecasting setting.  Specifically, each forecasting instance consists of two heterogeneous sources of information: numerical time-series observations and textual context. 
The numerical modality corresponds to the look-back window, which is partitioned into continuous patches and projected into the model hidden space.  The textual modality corresponds to context features, including domain knowledge, task instructions, and statistical information, which are processed by the native tokenizer and embedding layer of the pre-trained LLM. 
These two modalities are concatenated and encoded jointly by the bidirectional encoder, allowing the model to condition future generation on both temporal patterns and semantic context. All textual information is designed solely from the historical look-back window and does not incorporate any future or target information, ensuring that no information leakage occurs.

As shown  in Table~\ref{tab:ablation_prompt}, to evaluate the specific contributions of various textual context components, we conducted a comprehensive ablation study across ten benchmarks by selectively removing domain knowledge, task instruction, and statistics information. As illustrated in the results, the full text-conditioned framework consistently achieves the best performance across all datasets and prediction horizons. Specifically, removing the entire textual input (w/o Text) leads to a universal degradation in forecasting precision, confirming that semantic context provides essential guidance for the generative process. Among the individual sub-components, the exclusion of statistics information (w/o Statistics) frequently results in a more pronounced increase in error compared to omitting domain features or instructions, as seen in datasets like Energy and Exchange. This suggests that explicit statistical metadata—such as mean and variance—is critical for the model to effectively align its generative distribution with the global data scale. Furthermore, the performance drop observed in the w/o Domain and w/o Instruction variants validates that high-level task descriptions and domain-specific knowledge further refine the model's understanding capability, enabling more accurate capture of diverse temporal patterns. These results empirically verify that a multi-faceted semantic conditioning strategy is vital for high-precision  time series forecasting.

\section{Comparison with Generative Models}

As shown  in Table~\ref{tab:generative_comparison_wide}, to verify the architectural advantages of our framework, we conducted a comprehensive comparison against two mainstream generative paradigms: vanilla flow-matching and diffusion models. As evidenced by the results across ten diverse benchmarks, our method consistently secures superior forecasting performance across all prediction horizons. Specifically, our model achieves a marked reduction in error metrics compared to Diffusion-based approaches, which often suffer from error accumulation during iterative denoising steps. Furthermore, our approach outperforms standard Flow-matching baselines by a significant margin, particularly on datasets with high volatility. This performance gap highlights the efficacy of our modeling objective, which explicitly targets a linearized generative path, thereby stabilizing the generation process and effectively bridging high-level semantic understanding with continuous probabilistic modeling.

As shown  in Table~\ref{tab:nfe_ablation_wide}, complementing this comparative analysis, we further examined the computational efficiency and stability of our generative process through an ablation study on the number of function evaluations (NFE). While traditional generative models typically require multiple sampling steps to ensure output fidelity, our results demonstrate that our framework achieves optimal or near-optimal forecasting accuracy with just a single evaluation step. Increasing the sampling steps to two or three yields only marginal performance gains in specific cases, while the one-step inference remains highly accurate across all datasets. This finding empirically confirms the straightness of the learned flow trajectories, proving that our linearized coupling strategy enables rapid, high-precision inference without the latency and computational overhead inherent in multi-step iterative solvers.

\section{Hyperparameter Sensitivity}
\label{sec:Hyper}
\subsection{Patch Size} 
As shown in Table~\ref{tab:patchsize_ablation}, to investigate the impact of local temporal granularity on forecasting performance, we conducted a sensitivity analysis by varying the patch size within the range of $\{2, 4, 6, 8, 10\}$. 
We adopt non-overlapping patches throughout this analysis. 
When the sequence length is not divisible by the patch size, we pad the sequence at the end to the nearest multiple of the patch size before patchification. 
After forecasting, we retain only the outputs corresponding to the original prediction length and discard the padded positions. 
This strategy allows patch sizes such as 10 to be evaluated under the fixed look-back length $L=96$ and prediction lengths $H \in \{12,24,36,48\}$ while keeping the non-overlapping patching protocol unchanged. 
As shown in the results, CoGenCast exhibits dataset-dependent sensitivity to the patch size. 
Intermediate patch sizes, especially 4 and 6, generally achieve the best balance between local temporal details and global structure. 
Smaller patches are more effective for datasets such as Environment and Exchange, where high-resolution tokenization helps capture fine-grained fluctuations. 
In contrast, overly large patches often degrade performance, particularly on Health, because they may smooth out short-term dynamics. 
Meanwhile, excessively small patches can be suboptimal for complex datasets such as Health and Wind due to insufficient semantic context within each token. 
These results indicate that an appropriate patch size is important for constructing meaningful temporal tokens.

\section{Visualization}
\label{sec:visulaiztion}
\begin{figure}[h]
    \centering
    \includegraphics[width=0.99\columnwidth]{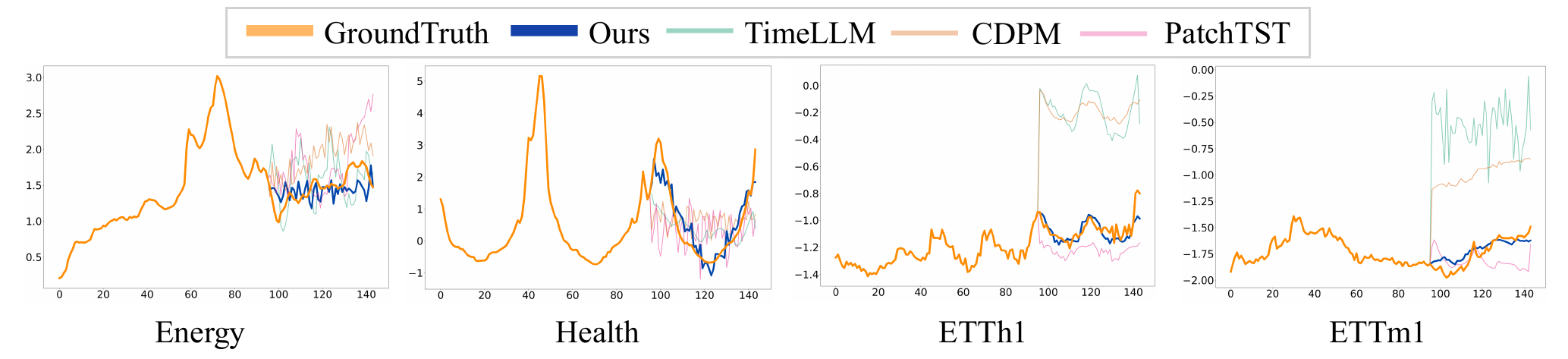}
    \caption{Comparison of forecasting results on Energy, Health, ETTh1, and ETTm1 datasets. We visualize the predicted trajectories of our method against the ground truth and leading baselines (Time-LLM, CDPM, and PatchTST).}
    \label{fig:visulization}
\end{figure}
To provide a qualitative assessment of forecasting fidelity, we visualize the predicted trajectories of our framework alongside the ground truth and competitive baselines, including Time-LLM, CDPM, and PatchTST. As illustrated in Figure \ref{fig:visulization}, our model consistently generates future sequences that align most closely with the ground truth across diverse datasets such as Energy, Health, and the ETT series. Specifically, in datasets characterized by high volatility and complex seasonal patterns like Health, our model effectively captures the underlying stochastic dynamics without the severe over-smoothing or spurious fluctuations observed in baseline methods. Furthermore, for the ETTh1 and ETTm1 benchmarks, while competing methods often exhibit significant distributional shifts or phase delays, our approach maintains a stable and precise trend. This demonstrates the capability of our hybrid generative framework to bridge global semantic understanding with local continuous refinement. These visualized results empirically confirm that our model not only achieves superior numerical performance but also produces reliable and physically meaningful forecasts for real-world time series applications.

\section{Limitations}

Although CoGenCast demonstrates strong forecasting performance across diverse benchmarks, several limitations remain. 
First, despite using a relatively compact LLM backbone and efficient one-step flow-based generation, CoGenCast still introduces additional computational costs compared with lightweight forecasting models, which may limit its deployment in resource-constrained scenarios. 
Second, the autoregressive appending strategy may suffer from error accumulation under extremely long prediction horizons. 
Third, our evaluation is mainly conducted on public benchmark datasets, and further validation is needed in real-world scenarios involving complex external factors, distribution shifts, missing observations, and deployment constraints. 
Finally, the multimodal setting in this work mainly refers to a practical numerical-textual forecasting scenario, where time-series observations are combined with textual contextual information. 
Future work will focus on reducing computational overhead, improving long-horizon generation stability, validating CoGenCast in broader real-world applications, and extending it to richer multimodal forecasting settings.

\section{Detailed Tables}
\label{sec: tables}
This section provides the full experimental results for all evaluated benchmarks to offer a more granular perspective on the performance of CoGenCast. These detailed tables present the comprehensive forecasting metrics, including Mean Squared Error (MSE) and Mean Absolute Error (MAE), across various datasets and prediction horizons. Furthermore, this section includes complete results from our ablation studies, exhaustive comparisons with generative models, and systematic sensitivity analyses of key hyperparameters. Additional extended experimental analyses are also provided to further demonstrate the effectiveness and scalability of our proposed framework across diverse scenarios.

\begin{table*}[h]
\centering
\caption{Full results of time series forecasting across all benchmark datasets. We provide a detailed comparison between our proposed framework and competitive baselines, including LLM-based, generative-based, and deep-based methods, evaluated on look-back window $L = 96$ and four predicted windows $ H \in \{12, 24, 36, 48\}$. The best results are in \textbf{bold} and the second best are \underline{underlined}.}
\label{tab:full_main_result}
\setlength{\tabcolsep}{3pt}
\renewcommand{\arraystretch}{1.25}
\small
\resizebox{\textwidth}{!}{%
\begin{tabular}{c|c|cc|cccccc|cccccc|cccccccccccccc}
\toprule

\multicolumn{1}{c}{\multirow{2}{*}{\centering \textsc{Methods}}} 
& \multicolumn{1}{c|}{} 
& \multicolumn{2}{c|}{\textsc{Ours}} 
& \multicolumn{6}{c|}{\textsc{LLM-based}} 
& \multicolumn{6}{c|}{\textsc{Generative-based}} 
& \multicolumn{14}{c}{\textsc{Deep-based}} \\[2pt]

\multicolumn{1}{c}{} 
& \multicolumn{1}{c|}{} 
& \multicolumn{2}{c|}{\textbf{CoGenCast}} 
& \multicolumn{2}{c}{TokenCast} 
& \multicolumn{2}{c}{TimeLLM} 
& \multicolumn{2}{c|}{LLM4TS} 
& \multicolumn{2}{c}{FlowTS} 
& \multicolumn{2}{c}{CDPM} 
& \multicolumn{2}{c|}{TimeGrad} 
& \multicolumn{2}{c}{ConvTimeNet} 
& \multicolumn{2}{c}{TimeDART} 
& \multicolumn{2}{c}{iTransformer} 
& \multicolumn{2}{c}{TimesNet} 
& \multicolumn{2}{c}{PatchTST} 
& \multicolumn{2}{c}{DLinear} 
& \multicolumn{2}{c}{Autoformer} \\[2pt]

\multicolumn{1}{c}{\centering \textsc{Metrics}} 
& \multicolumn{1}{c|}{} 
& MSE & MAE 
& MSE & MAE & MSE & MAE & MSE & MAE
& MSE & MAE & MSE & MAE & MSE & MAE
& MSE & MAE & MSE & MAE & MSE & MAE & MSE & MAE & MSE & MAE & MSE & MAE & MSE & MAE \\
\midrule

\multirow{4}{*}{Energy}
& 12 & \textbf{0.140} & \textbf{0.273} & 0.175 & 0.318 & 0.177 & 0.321 & 0.180 & 0.324 & 0.354 & 0.402 & 0.210 & 0.351 & 0.612 & 0.609 & \underline{0.168} & \underline{0.311} & 0.177 & 0.321 & 0.171 & 0.316 & 0.176 & 0.322 & 0.175 & 0.320 & 0.179 & 0.326 & 0.461 & 0.518 \\
& 24 & \textbf{0.265} & \textbf{0.379} & 0.289 & 0.401 & 0.293 & 0.405 & 0.291 & 0.404 & 0.392 & 0.448 & 0.349 & 0.427 & 0.649 & 0.627 & \underline{0.279} & \underline{0.391} & 0.283 & 0.394 & 0.285 & 0.406 & 0.291 & 0.402 & 0.288 & 0.398 & 0.295 & 0.405 & 0.484 & 0.528 \\
& 36 & \textbf{0.327} & \textbf{0.428} & \underline{0.339} & 0.443 & 0.340 & \underline{0.442} & 0.345 & 0.445 & 0.441 & 0.503 & 0.460 & 0.484 & 0.696 & 0.642 & 0.356 & 0.451 & 0.362 & 0.455 & 0.374 & 0.464 & 0.371 & 0.459 & 0.366 & 0.457 & 0.377 & 0.467 & 0.488 & 0.540 \\
& 48 & \textbf{0.441} & 0.499 & 0.462 & \underline{0.485} & \underline{0.461} & \textbf{0.483} & 0.482 & 0.521 & 0.485 & 0.539 & 0.473 & 0.524 & 0.738 & 0.658 & 0.465 & 0.510 & 0.469 & 0.513 & 0.475 & 0.523 & 0.483 & 0.525 & 0.479 & 0.522 & 0.487 & 0.527 & 0.591 & 0.581 \\
\midrule

\multirow{4}{*}{ETTh1}
& 12 & \textbf{0.290} & \textbf{0.343} & 0.312 & 0.364 & 0.318 & 0.364 & 0.314 & 0.365 & \underline{0.311} & \underline{0.352} & 0.313 & 0.359 & 0.832 & 0.742 & 0.315 & 0.359 & 0.319 & 0.362 & 0.323 & 0.366 & 0.324 & 0.367 & 0.321 & 0.365 & 0.319 & 0.362 & 0.434 & 0.416 \\
& 24 & \textbf{0.319} & \textbf{0.362} & 0.331 & 0.377 & 0.387 & 0.406 & \underline{0.330} & 0.375 & 0.356 & 0.389 & 0.337 & \underline{0.374} & 0.874 & 0.763 & 0.345 & 0.379 & 0.343 & 0.376 & 0.350 & 0.383 & 0.349 & 0.381 & 0.352 & 0.382 & 0.348 & 0.379 & 0.481 & 0.470 \\
& 36 & \textbf{0.344} & \textbf{0.377} & \underline{0.355} & \underline{0.384} & 0.358 & 0.387 & 0.358 & 0.385 & 0.404 & 0.431 & 0.356 & 0.387 & 0.912 & 0.789 & 0.362 & 0.389 & 0.366 & 0.391 & 0.371 & 0.394 & 0.378 & 0.399 & 0.384 & 0.401 & 0.382 & 0.402 & 0.519 & 0.492 \\
& 48 & \textbf{0.357} & \textbf{0.386} & \underline{0.382} & \underline{0.399} & 0.393 & 0.408 & 0.388 & 0.402 & 0.453 & 0.480 & 0.392 & 0.410 & 0.953 & 0.810 & \underline{0.382} & 0.401 & 0.389 & 0.405 & 0.385 & 0.407 & 0.392 & 0.409 & 0.411 & 0.413 & 0.401 & 0.408 & 0.544 & 0.498 \\
\midrule

\multirow{4}{*}{ETTh2}
& 12 & 0.121 & \textbf{0.216} & 0.133 & 0.235 & 0.146 & 0.249 & 0.135 & 0.236 & 0.168 & 0.261 & 0.145 & 0.248 & 0.534 & 0.508 & 0.125 & 0.231 & \underline{0.111} & 0.226 & 0.140 & 0.243 & 0.126 & 0.234 & \textbf{0.110} & \underline{0.225} & 0.128 & 0.236 & 0.172 & 0.283 \\
& 24 & 0.152 & \textbf{0.243} & 0.187 & 0.280 & 0.199 & 0.289 & 0.190 & 0.282 & 0.192 & 0.283 & 0.175 & 0.267 & 0.561 & 0.528 & 0.159 & 0.258 & \underline{0.133} & \underline{0.246} & 0.181 & 0.272 & 0.161 & 0.262 & \textbf{0.129} & \textbf{0.243} & 0.159 & 0.257 & 0.190 & 0.302 \\
& 36 & 0.172 & \textbf{0.254} & 0.216 & 0.298 & 0.232 & 0.309 & 0.221 & 0.302 & 0.249 & 0.318 & 0.222 & 0.300 & 0.583 & 0.544 & 0.178 & 0.278 & \underline{0.155} & 0.263 & 0.216 & 0.297 & 0.175 & 0.276 & \textbf{0.150} & \underline{0.259} & 0.177 & 0.279 & 0.197 & 0.308 \\
& 48 & 0.191 & \textbf{0.272} & 0.227 & 0.301 & 0.246 & 0.317 & 0.234 & 0.305 & 0.263 & 0.339 & 0.229 & 0.301 & 0.607 & 0.582 & 0.199 & 0.297 & \underline{0.179} & 0.278 & 0.240 & 0.312 & 0.194 & 0.293 & \textbf{0.178} & \underline{0.276} & 0.195 & 0.293 & 0.209 & 0.312 \\
\midrule

\multirow{4}{*}{ETTm1}
& 12 & \textbf{0.140} & \textbf{0.232} & 0.167 & 0.258 & 0.182 & 0.271 & 0.173 & 0.262 & 0.267 & 0.320 & 0.201 & 0.304 & 0.714 & 0.578 & \underline{0.149} & \underline{0.241} & 0.159 & 0.262 & 0.151 & 0.244 & 0.162 & 0.265 & 0.165 & 0.267 & 0.168 & 0.269 & 0.282 & 0.347 \\
& 24 & \textbf{0.213} & \textbf{0.280} & 0.245 & 0.311 & 0.323 & 0.365 & 0.248 & 0.312 & 0.305 & 0.351 & 0.309 & 0.351 & 0.782 & 0.603 & \underline{0.230} & \underline{0.299} & 0.231 & 0.303 & 0.234 & 0.304 & 0.234 & 0.305 & 0.236 & 0.308 & 0.238 & 0.311 & 0.337 & 0.391 \\
& 36 & \textbf{0.255} & \textbf{0.306} & 0.281 & 0.331 & 0.361 & 0.381 & 0.285 & 0.334 & 0.344 & 0.391 & 0.333 & 0.369 & 0.824 & 0.629 & 0.276 & 0.331 & 0.274 & 0.328 & 0.281 & 0.335 & 0.274 & 0.329 & \underline{0.271} & \underline{0.327} & 0.275 & 0.329 & 0.399 & 0.442 \\
& 48 & \textbf{0.288} & \textbf{0.321} & 0.301 & \underline{0.342} & 0.349 & 0.377 & 0.307 & 0.346 & 0.392 & 0.434 & 0.365 & 0.382 & 0.884 & 0.647 & 0.302 & 0.345 & \underline{0.297} & 0.343 & 0.305 & 0.349 & 0.309 & 0.348 & 0.301 & 0.345 & 0.310 & 0.352 & 0.483 & 0.549 \\
\midrule

\multirow{4}{*}{ETTm2}
& 12 & \textbf{0.079} & \textbf{0.167} & 0.111 & 0.219 & 0.109 & 0.218 & 0.115 & 0.227 & 0.149 & 0.256 & 0.144 & 0.252 & 0.598 & 0.429 & 0.103 & 0.212 & \underline{0.101} & \underline{0.201} & \underline{0.101} & 0.207 & 0.103 & 0.210 & 0.102 & 0.203 & 0.105 & 0.214 & 0.218 & 0.347 \\
& 24 & \textbf{0.102} & \textbf{0.194} & 0.137 & 0.244 & 0.139 & 0.245 & 0.138 & 0.247 & 0.191 & 0.299 & 0.185 & 0.296 & 0.652 & 0.467 & 0.114 & \underline{0.225} & \underline{0.113} & \underline{0.225} & 0.114 & 0.229 & 0.116 & 0.231 & 0.117 & 0.228 & 0.119 & 0.235 & 0.234 & 0.365 \\
& 36 & \textbf{0.120} & \textbf{0.212} & 0.151 & 0.255 & 0.153 & 0.256 & 0.156 & 0.262 & 0.216 & 0.326 & 0.210 & 0.321 & 0.763 & 0.529 & 0.135 & 0.244 & \underline{0.130} & \underline{0.239} & 0.132 & 0.243 & 0.134 & 0.247 & 0.134 & 0.241 & 0.137 & 0.249 & 0.246 & 0.372 \\
& 48 & \textbf{0.134} & \textbf{0.226} & 0.166 & 0.269 & 0.162 & 0.268 & 0.170 & 0.271 & 0.233 & 0.339 & 0.225 & 0.333 & 0.826 & 0.614 & 0.152 & 0.262 & \underline{0.146} & \underline{0.261} & 0.149 & 0.265 & 0.151 & 0.268 & 0.151 & 0.266 & 0.154 & 0.270 & 0.254 & 0.376 \\
\midrule

\multirow{4}{*}{Environ.}
& 12 & \textbf{0.286} & \underline{0.372} & 0.297 & 0.385 & 0.298 & 0.386 & 0.302 & 0.388 & \textbf{0.286} & \textbf{0.361} & 0.319 & 0.428 & 0.781 & 0.625 & 0.293 & 0.382 & 0.295 & 0.389 & 0.296 & 0.393 & 0.298 & 0.395 & \underline{0.292} & 0.387 & 0.293 & 0.384 & 0.346 & 0.436 \\
& 24 & \textbf{0.295} & \textbf{0.378} & 0.316 & 0.399 & 0.319 & 0.401 & 0.312 & \underline{0.393} & 0.322 & 0.404 & 0.330 & 0.397 & 0.828 & 0.669 & 0.312 & 0.399 & 0.307 & 0.395 & 0.306 & 0.398 & 0.305 & 0.398 & \underline{0.302} & \underline{0.393} & 0.310 & 0.397 & 0.358 & 0.459 \\
& 36 & \textbf{0.305} & \textbf{0.383} & 0.319 & \underline{0.395} & 0.325 & 0.403 & 0.321 & 0.398 & 0.369 & 0.447 & 0.349 & 0.444 & 0.854 & 0.693 & 0.326 & 0.402 & 0.321 & 0.404 & 0.321 & 0.406 & 0.323 & 0.407 & \underline{0.315} & 0.400 & 0.320 & 0.405 & 0.379 & 0.475 \\
& 48 & \textbf{0.316} & \textbf{0.390} & 0.331 & 0.403 & 0.337 & 0.411 & 0.330 & \underline{0.401} & 0.419 & 0.488 & 0.338 & 0.407 & 0.912 & 0.732 & 0.332 & 0.407 & 0.333 & 0.409 & 0.335 & 0.412 & 0.336 & 0.413 & \underline{0.327} & 0.407 & 0.332 & 0.409 & 0.381 & 0.473 \\
\midrule

\multirow{4}{*}{Exchange}
& 12 & \textbf{0.015} & \textbf{0.079} & 0.021 & 0.102 & 0.019 & 0.096 & 0.023 & 0.106 & 0.041 & 0.138 & 0.029 & 0.124 & 0.237 & 0.398 & \underline{0.017} & 0.095 & \underline{0.017} & 0.094 & 0.018 & 0.095 & 0.018 & 0.097 & 0.018 & \underline{0.093} & 0.019 & 0.098 & 0.057 & 0.173 \\
& 24 & \textbf{0.026} & \textbf{0.110} & 0.032 & 0.124 & 0.030 & 0.121 & 0.032 & 0.125 & 0.054 & 0.169 & 0.034 & 0.131 & 0.291 & 0.432 & 0.031 & 0.122 & 0.030 & 0.119 & 0.031 & 0.121 & 0.030 & 0.120 & \underline{0.028} & \underline{0.118} & 0.031 & 0.124 & 0.084 & 0.212 \\
& 36 & \textbf{0.037} & \textbf{0.132} & 0.040 & 0.143 & \underline{0.039} & \underline{0.141} & 0.041 & 0.142 & 0.069 & 0.201 & 0.057 & 0.173 & 0.356 & 0.465 & 0.041 & 0.143 & 0.040 & 0.142 & 0.040 & 0.143 & 0.041 & 0.144 & \underline{0.039} & 0.143 & 0.042 & 0.145 & 0.089 & 0.219 \\
& 48 & \textbf{0.046} & \textbf{0.150} & 0.054 & 0.165 & 0.053 & 0.163 & 0.051 & 0.158 & 0.080 & 0.232 & 0.065 & 0.183 & 0.394 & 0.482 & \underline{0.048} & 0.154 & \underline{0.048} & \underline{0.153} & 0.051 & 0.159 & 0.052 & 0.161 & \underline{0.048} & 0.155 & 0.054 & 0.164 & 0.117 & 0.250 \\
\midrule

\multirow{4}{*}{Health}
& 12 & \textbf{1.056} & \textbf{0.697} & 1.302 & 0.778 & 1.311 & 0.781 & 1.351 & 0.809 & 1.482 & 0.824 & 1.298 & 0.788 & 2.174 & 1.232 & 1.187 & 0.741 & \underline{1.119} & \underline{0.736} & 1.142 & 0.752 & 1.204 & 0.771 & 1.224 & 0.750 & 1.235 & 0.759 & 1.737 & 0.987 \\
& 24 & \textbf{1.500} & \textbf{0.820} & 1.562 & 0.868 & 1.602 & 0.865 & 1.570 & 0.875 & 1.621 & 0.901 & 1.569 & 0.870 & 2.342 & 1.387 & 1.591 & 0.879 & \underline{1.544} & \underline{0.851} & 1.587 & 0.869 & 1.552 & 0.854 & 1.556 & 0.857 & 1.549 & 0.859 & 1.809 & 1.014 \\
& 36 & \textbf{1.583} & \textbf{0.852} & 1.698 & 0.897 & 1.713 & 0.901 & 1.878 & 1.001 & 1.815 & 0.985 & 1.682 & 0.907 & 2.553 & 1.492 & 1.652 & 0.892 & 1.639 & 0.892 & 1.669 & 0.919 & 1.727 & 0.910 & \underline{1.636} & \underline{0.889} & 1.670 & 0.913 & 1.845 & 1.018 \\
& 48 & \textbf{1.638} & \textbf{0.879} & 1.742 & 0.963 & 1.818 & 0.999 & 1.918 & 1.007 & 2.026 & 1.062 & 1.815 & 1.022 & 2.874 & 1.644 & 1.739 & 0.932 & 1.737 & \underline{0.927} & \underline{1.736} & 0.945 & 1.842 & 0.930 & 1.743 & 0.931 & 1.829 & 0.938 & 1.919 & 1.051 \\
\midrule

\multirow{4}{*}{Wind}
& 12 & \textbf{0.269} & \textbf{0.263} & 0.283 & 0.288 & 0.281 & 0.289 & 0.351 & 0.355 & 0.521 & 0.424 & 0.350 & 0.342 & 0.805 & 0.612 & 0.279 & 0.288 & 0.278 & 0.289 & 0.282 & 0.291 & 0.281 & 0.291 & \underline{0.274} & \underline{0.286} & 0.278 & 0.288 & 0.647 & 0.532 \\
& 24 & \textbf{0.479} & \textbf{0.399} & 0.485 & \underline{0.415} & 0.486 & 0.417 & 0.536 & 0.457 & 0.603 & 0.467 & 0.533 & 0.444 & 0.913 & 0.665 & 0.487 & 0.417 & 0.486 & 0.417 & 0.490 & 0.421 & 0.488 & 0.420 & \underline{0.484} & \underline{0.415} & 0.489 & 0.421 & 0.765 & 0.584 \\
& 36 & \textbf{0.637} & \textbf{0.488} & 0.652 & 0.511 & 0.655 & 0.512 & 0.675 & 0.526 & 0.692 & 0.518 & 0.667 & 0.514 & 0.968 & 0.677 & 0.653 & 0.510 & 0.652 & 0.508 & 0.657 & 0.516 & 0.656 & 0.514 & \underline{0.649} & \underline{0.504} & 0.655 & 0.513 & 0.857 & 0.618 \\
& 48 & \textbf{0.741} & \textbf{0.549} & 0.755 & 0.562 & 0.751 & 0.559 & 0.801 & 0.599 & 0.788 & 0.579 & 0.782 & 0.571 & 1.129 & 0.711 & 0.756 & 0.560 & 0.750 & 0.559 & 0.758 & 0.568 & 0.760 & 0.565 & \underline{0.749} & \underline{0.557} & 0.757 & 0.565 & 1.051 & 0.701 \\
\midrule

\multirow{4}{*}{Solar}
& 12 & \textbf{0.115} & \textbf{0.206} & \underline{0.121} & \underline{0.211} & 0.122 & \underline{0.211} & 0.124 & 0.222 & 0.295 & 0.341 & 0.237 & 0.340 & 0.782 & 0.687 & 0.123 & 0.214 & 0.123 & 0.213 & 0.131 & 0.225 & 0.127 & 0.218 & \underline{0.121} & 0.212 & 0.125 & 0.216 & 0.407 & 0.507 \\
& 24 & \textbf{0.216} & \textbf{0.285} & 0.226 & 0.295 & 0.228 & 0.298 & 0.236 & 0.309 & 0.344 & 0.396 & 0.352 & 0.414 & 0.854 & 0.723 & 0.227 & 0.297 & 0.228 & 0.297 & 0.236 & 0.309 & 0.231 & 0.301 & \underline{0.223} & \underline{0.294} & 0.229 & 0.298 & 0.725 & 0.671 \\
& 36 & \textbf{0.288} & \textbf{0.338} & 0.291 & 0.341 & \underline{0.289} & \underline{0.339} & 0.293 & 0.348 & 0.398 & 0.453 & 0.407 & 0.451 & 0.972 & 0.785 & 0.302 & 0.352 & 0.301 & 0.351 & 0.315 & 0.362 & 0.305 & 0.354 & 0.297 & 0.347 & 0.304 & 0.353 & 0.594 & 0.586 \\
& 48 & \textbf{0.276} & \textbf{0.328} & 0.295 & 0.345 & 0.294 & 0.346 & 0.301 & 0.349 & 0.439 & 0.498 & 0.405 & 0.436 & 1.036 & 0.861 & 0.292 & 0.345 & 0.291 & 0.343 & 0.302 & 0.357 & 0.295 & 0.346 & \underline{0.286} & \underline{0.339} & 0.292 & 0.344 & 0.917 & 0.741 \\

\bottomrule
\end{tabular}%
}
\end{table*}

\begin{table*}[h]
\centering
\caption{ Full results of in-domain and cross-domain training across all benchmark datasets. Results are reported in MSE and MAE for each predicted windows $ H \in \{12, 24, 36, 48\}$. The best results are in \textbf{bold}.}
\label{tab:full_cross_domain_result}
\setlength{\tabcolsep}{2pt} 
\renewcommand{\arraystretch}{1.1}
\small
\resizebox{\textwidth}{!}{%
\begin{tabular}{c c | cc | cc | cc | cc | cc | cc | cc | cc | cc | cc}
\toprule
\multirow{2}{*}{\textsc{Horizon}} & \multirow{2}{*}{\textsc{Metrics}} & \multicolumn{2}{c|}{\textsc{Energy}} & \multicolumn{2}{c|}{\textsc{ETTh1}} & \multicolumn{2}{c|}{\textsc{ETTh2}} & \multicolumn{2}{c|}{\textsc{ETTm1}} & \multicolumn{2}{c|}{\textsc{ETTm2}} & \multicolumn{2}{c|}{\textsc{Environ.}} & \multicolumn{2}{c|}{\textsc{Exchange}} & \multicolumn{2}{c|}{\textsc{Health}} & \multicolumn{2}{c|}{\textsc{Wind}} & \multicolumn{2}{c}{\textsc{Solar}} \\
& & In & Cross & In & Cross & In & Cross & In & Cross & In & Cross & In & Cross & In & Cross & In & Cross & In & Cross & In & Cross \\
\midrule
\multirow{2}{*}{12} & MSE & 0.140 & \textbf{0.137} & 0.290 & \textbf{0.286} & 0.121 & \textbf{0.119} & 0.140 & \textbf{0.138} & 0.079 & \textbf{0.078} & 0.286 & \textbf{0.282} & \textbf{0.015} & \textbf{0.015} & 1.056 & \textbf{1.049} & 0.269 & \textbf{0.267} & 0.115 & \textbf{0.114} \\
& MAE & 0.273 & \textbf{0.271} & 0.343 & \textbf{0.341} & 0.216 & \textbf{0.215} & 0.232 & \textbf{0.230} & 0.167 & \textbf{0.166} & 0.372 & \textbf{0.368} & \textbf{0.079} & \textbf{0.079} & 0.697 & \textbf{0.692} & 0.263 & \textbf{0.262} & \textbf{0.206} & \textbf{0.206} \\
\cmidrule{1-22}
\multirow{2}{*}{24} & MSE & 0.265 & \textbf{0.262} & 0.319 & \textbf{0.315} & 0.152 & \textbf{0.151} & 0.213 & \textbf{0.210} & 0.102 & \textbf{0.100} & 0.295 & \textbf{0.291} & 0.026 & \textbf{0.025} & 1.500 & \textbf{1.487} & 0.479 & \textbf{0.475} & 0.216 & \textbf{0.214} \\
& MAE & 0.379 & \textbf{0.375} & 0.362 & \textbf{0.359} & 0.243 & \textbf{0.242} & 0.280 & \textbf{0.278} & 0.194 & \textbf{0.191} & 0.378 & \textbf{0.375} & \textbf{0.110} & \textbf{0.110} & 0.820 & \textbf{0.812} & 0.399 & \textbf{0.397} & 0.285 & \textbf{0.283} \\
\cmidrule{1-22}
\multirow{2}{*}{36} & MSE & 0.327 & \textbf{0.323} & 0.344 & \textbf{0.341} & 0.172 & \textbf{0.170} & 0.255 & \textbf{0.252} & 0.120 & \textbf{0.116} & 0.305 & \textbf{0.302} & 0.037 & \textbf{0.036} & 1.583 & \textbf{1.539} & 0.637 & \textbf{0.629} & 0.288 & \textbf{0.284} \\
& MAE & 0.428 & \textbf{0.424} & 0.377 & \textbf{0.373} & 0.254 & \textbf{0.251} & 0.306 & \textbf{0.304} & 0.212 & \textbf{0.209} & 0.383 & \textbf{0.379} & 0.132 & \textbf{0.130} & 0.852 & \textbf{0.846} & 0.488 & \textbf{0.483} & 0.338 & \textbf{0.335} \\
\cmidrule{1-22}
\multirow{2}{*}{48} & MSE & 0.441 & \textbf{0.436} & 0.357 & \textbf{0.352} & 0.191 & \textbf{0.189} & 0.288 & \textbf{0.285} & 0.134 & \textbf{0.131} & 0.316 & \textbf{0.311} & 0.046 & \textbf{0.044} & 1.638 & \textbf{1.614} & 0.741 & \textbf{0.736} & 0.276 & \textbf{0.273} \\
& MAE & 0.499 & \textbf{0.491} & \textbf{0.386} & 0.389 & 0.272 & \textbf{0.271} & 0.321 & \textbf{0.319} & 0.226 & \textbf{0.222} & 0.390 & \textbf{0.386} & 0.150 & \textbf{0.147} & 0.879 & \textbf{0.865} & \textbf{0.549} & 0.550 & 0.328 & \textbf{0.325} \\
\bottomrule
\end{tabular}%
}
\end{table*}

\begin{table*}[h]
\centering
\caption{Quantitative results of cross-domain few-shot forecasting. We evaluate the performance across representative datasets under different training data scales ($5\%, 10\%, 20\%, 100\%$) to demonstrate the data efficiency. The best results are in \textbf{bold}.}
\label{tab:cross_domain_fewshot}

\scriptsize
\setlength{\tabcolsep}{5pt}
\renewcommand{\arraystretch}{0.8}

\begin{tabular}{l|c|cc|cc|cc|cc}
\toprule

\multicolumn{2}{c|}{\textsc{Portion}} 
& \multicolumn{2}{c|}{\textbf{5\%}} 
& \multicolumn{2}{c|}{\textbf{10\%}} 
& \multicolumn{2}{c|}{\textbf{20\%}} 
& \multicolumn{2}{c}{\textbf{Full (100\%)}} \\

\multicolumn{2}{c|}{\textsc{Metrics}} 
& MSE & MAE 
& MSE & MAE 
& MSE & MAE 
& MSE & MAE \\

\midrule

\multirow{4}{*}{ETTh1}
& 12  & 0.813 & 0.591 & 0.498 & 0.463 & 0.340 & 0.378 & \textbf{0.290} & \textbf{0.343} \\
& 24  & 0.871 & 0.613 & 0.536 & 0.483 & 0.378 & 0.400 & \textbf{0.319} & \textbf{0.362} \\
& 36  & 0.924 & 0.634 & 0.560 & 0.495 & 0.408 & 0.418 & \textbf{0.344} & \textbf{0.377} \\
& 48  & 0.941 & 0.642 & 0.576 & 0.503 & 0.428 & 0.427 & \textbf{0.357} & \textbf{0.386} \\
\cmidrule{1-10}

\multirow{4}{*}{ETTh2}
& 12  & 0.261 & 0.340 & 0.161 & 0.258 & 0.137 & 0.232 & \textbf{0.121} & \textbf{0.216} \\
& 24  & 0.302 & 0.361 & 0.195 & 0.281 & 0.171 & 0.261 & \textbf{0.152} & \textbf{0.243} \\
& 36  & 0.331 & 0.378 & 0.226 & 0.301 & 0.207 & 0.282 & \textbf{0.172} & \textbf{0.254} \\
& 48  & 0.348 & 0.389 & 0.251 & 0.316 & 0.228 & 0.294 & \textbf{0.191} & \textbf{0.272} \\
\cmidrule{1-10}

\multirow{4}{*}{ETTm1}
& 12  & 0.191 & 0.273 & 0.189 & 0.270 & 0.161 & 0.251 & \textbf{0.140} & \textbf{0.232} \\
& 24  & 0.292 & 0.336 & 0.306 & 0.335 & 0.247 & 0.304 & \textbf{0.213} & \textbf{0.280} \\
& 36  & 0.353 & 0.368 & 0.377 & 0.372 & 0.293 & 0.332 & \textbf{0.255} & \textbf{0.306} \\
& 48  & 0.386 & 0.385 & 0.421 & 0.391 & 0.314 & 0.347 & \textbf{0.288} & \textbf{0.321} \\
\cmidrule{1-10}

\multirow{4}{*}{ETTm2}
& 12  & 0.092 & 0.191 & 0.084 & 0.178 & 0.081 & 0.175 & \textbf{0.079} & \textbf{0.167} \\
& 24  & 0.129 & 0.224 & 0.114 & 0.209 & 0.111 & 0.208 & \textbf{0.102} & \textbf{0.194} \\
& 36  & 0.154 & 0.247 & 0.137 & 0.230 & 0.132 & 0.228 & \textbf{0.120} & \textbf{0.212} \\
& 48  & 0.175 & 0.262 & 0.155 & 0.244 & 0.149 & 0.241 & \textbf{0.134} & \textbf{0.226} \\
\cmidrule{1-10}

\multirow{4}{*}{Environ.}
& 12  & 0.612 & 0.576 & 0.348 & 0.423 & 0.312 & 0.398 & \textbf{0.286} & \textbf{0.372} \\
& 24  & 0.626 & 0.582 & 0.363 & 0.436 & 0.320 & 0.406 & \textbf{0.295} & \textbf{0.378} \\
& 36  & 0.657 & 0.596 & 0.380 & 0.448 & 0.336 & 0.416 & \textbf{0.305} & \textbf{0.383} \\
& 48  & 0.676 & 0.604 & 0.392 & 0.457 & 0.349 & 0.426 & \textbf{0.316} & \textbf{0.390} \\
\cmidrule{1-10}

\multirow{4}{*}{Exchange}
& 12  & 0.026 & 0.108 & 0.025 & 0.108 & 0.019 & 0.092 & \textbf{0.015} & \textbf{0.079} \\
& 24  & 0.045 & 0.144 & 0.043 & 0.143 & 0.034 & 0.124 & \textbf{0.026} & \textbf{0.110} \\
& 36  & 0.062 & 0.170 & 0.059 & 0.167 & 0.043 & 0.143 & \textbf{0.037} & \textbf{0.132} \\
& 48  & 0.077 & 0.190 & 0.073 & 0.188 & 0.055 & 0.162 & \textbf{0.046} & \textbf{0.150} \\
\bottomrule

\end{tabular}
\end{table*}

\begin{table*}[h]
\centering
\caption{Full results of the ablation study on the encoder-decoder architecture. We compare our framework against encoder-only and decoder-only variants equipped with Qwen3-0.6B and Qwen3-1.7B backbones across all benchmark datasets. The best results are in \textbf{bold}.}
\label{tab:main_results_compact}

\scriptsize                         
\setlength{\tabcolsep}{5pt}     
\renewcommand{\arraystretch}{0.8} 

\begin{tabular}{l|c|cc|cccc|cccc}
\toprule

\multicolumn{2}{c|}{\textsc{Methods}} & \multicolumn{2}{c|}{\multirow{2}{*}{\textsc{Ours}}} & \multicolumn{4}{c|}{\textsc{Encoder-only}} & \multicolumn{4}{c}{\textsc{Decoder-only}} \\

\multicolumn{2}{c|}{} & \multicolumn{2}{c|}{} & \multicolumn{2}{c}{Qwen3-0.6B} & \multicolumn{2}{c|}{Qwen3-1.7B} & \multicolumn{2}{c}{Qwen3-0.6B} & \multicolumn{2}{c}{Qwen3-1.7B} \\

\multicolumn{2}{c|}{\textsc{Metrics}} & MSE & MAE & MSE & MAE & MSE & MAE & MSE & MAE & MSE & MAE \\ 
\midrule


\multirow{4}{*}{Energy} & 12 & \textbf{0.140} & \textbf{0.273} & 0.180 & 0.331 & 0.179 & 0.331 & 0.510 & 0.537 & 0.479 & 0.519 \\
 & 24 & \textbf{0.265} & \textbf{0.379} & 0.293 & 0.403 & 0.292 & 0.402 & 0.810 & 0.715 & 0.762 & 0.688 \\
 & 36 & \textbf{0.327} & \textbf{0.428} & 0.355 & 0.447 & 0.354 & 0.446 & 1.166 & 0.831 & 1.090 & 0.798 \\
 & 48 & \textbf{0.441} & \textbf{0.499} & 0.495 & 0.540 & 0.494 & 0.539 & 1.324 & 0.884 & 1.247 & 0.854 \\
\cmidrule{1-12}

\multirow{4}{*}{ETTh1} & 12 & \textbf{0.290} & \textbf{0.343} & 0.307 & 0.353 & 0.302 & 0.354 & 0.945 & 0.655 & 0.817 & 0.626 \\
 & 24 & \textbf{0.319} & \textbf{0.362} & 0.346 & 0.391 & 0.338 & 0.376 & 1.047 & 0.691 & 0.899 & 0.663 \\
 & 36 & \textbf{0.344} & \textbf{0.377} & 0.356 & 0.388 & 0.352 & 0.390 & 1.134 & 0.720 & 0.967 & 0.688 \\
 & 48 & \textbf{0.357} & \textbf{0.386} & 0.372 & 0.398 & 0.368 & 0.401 & 1.213 & 0.745 & 1.006 & 0.707 \\
\cmidrule{1-12}

\multirow{4}{*}{ETTh2} & 12 & \textbf{0.121} & \textbf{0.216} & 0.146 & 0.239 & 0.129 & 0.225 & 0.386 & 0.404 & 0.352 & 0.389 \\
 & 24 & \textbf{0.152} & \textbf{0.243} & 0.169 & 0.257 & 0.165 & 0.256 & 0.482 & 0.456 & 0.440 & 0.438 \\
 & 36 & \textbf{0.172} & \textbf{0.254} & 0.185 & 0.266 & 0.184 & 0.265 & 0.540 & 0.474 & 0.493 & 0.456 \\
 & 48 & \textbf{0.191} & \textbf{0.272} & 0.206 & 0.284 & 0.206 & 0.286 & 0.600 & 0.510 & 0.547 & 0.489 \\
\cmidrule{1-12}

\multirow{4}{*}{ETTm1} & 12 & \textbf{0.140} & \textbf{0.232} & 0.159 & 0.256 & 0.154 & 0.251 & 0.760 & 0.562 & 0.612 & 0.484 \\
 & 24 & \textbf{0.213} & \textbf{0.280} & 0.240 & 0.309 & 0.234 & 0.305 & 1.045 & 0.650 & 0.934 & 0.586 \\
 & 36 & \textbf{0.255} & \textbf{0.306} & 0.273 & 0.332 & 0.268 & 0.331 & 1.270 & 0.709 & 1.116 & 0.638 \\
 & 48 & \textbf{0.288} & \textbf{0.321} & 0.309 & 0.351 & 0.304 & 0.349 & 1.506 & 0.762 & 1.262 & 0.672 \\
\cmidrule{1-12}

\multirow{4}{*}{ETTm2} & 12 & \textbf{0.079} & \textbf{0.167} & 0.091 & 0.184 & 0.086 & 0.175 & 0.338 & 0.408 & 0.289 & 0.359 \\
 & 24 & \textbf{0.102} & \textbf{0.194} & 0.119 & 0.216 & 0.114 & 0.206 & 0.439 & 0.476 & 0.375 & 0.420 \\
 & 36 & \textbf{0.120} & \textbf{0.212} & 0.138 & 0.234 & 0.131 & 0.223 & 0.514 & 0.518 & 0.439 & 0.456 \\
 & 48 & \textbf{0.134} & \textbf{0.226} & 0.156 & 0.250 & 0.149 & 0.240 & 0.577 & 0.554 & 0.493 & 0.489 \\
\cmidrule{1-12}

\multirow{4}{*}{Environ.} & 12 & \textbf{0.286} & \textbf{0.372} & 0.298 & 0.388 & 0.296 & 0.393 & 0.576 & 0.565 & 0.536 & 0.539 \\
 & 24 & \textbf{0.295} & \textbf{0.378} & 0.303 & 0.402 & 0.302 & 0.401 & 0.590 & 0.568 & 0.552 & 0.548 \\
 & 36 & \textbf{0.305} & \textbf{0.383} & 0.317 & 0.409 & 0.315 & 0.404 & 0.613 & 0.576 & 0.574 & 0.555 \\
 & 48 & \textbf{0.316} & \textbf{0.390} & 0.333 & 0.415 & 0.329 & 0.414 & 0.629 & 0.586 & 0.589 & 0.567 \\
\cmidrule{1-12}

\multirow{4}{*}{Exchange} & 12 & \textbf{0.015} & \textbf{0.079} & 0.020 & 0.098 & 0.018 & 0.092 & 0.049 & 0.152 & 0.081 & 0.171 \\
 & 24 & \textbf{0.026} & \textbf{0.110} & 0.031 & 0.124 & 0.030 & 0.126 & 0.100 & 0.210 & 0.143 & 0.222 \\
 & 36 & \textbf{0.037} & \textbf{0.132} & 0.040 & 0.142 & 0.039 & 0.141 & 0.198 & 0.280 & 0.201 & 0.265 \\
 & 48 & \textbf{0.046} & \textbf{0.150} & 0.052 & 0.163 & 0.051 & 0.157 & 0.406 & 0.377 & 0.252 & 0.303 \\
\cmidrule{1-12}

\multirow{4}{*}{Health} & 12 & \textbf{1.056} & \textbf{0.697} & 1.165 & 0.714 & 1.155 & 0.740 & 3.891 & 1.528 & 2.622 & 1.193 \\
 & 24 & \textbf{1.500} & \textbf{0.820} & 1.624 & 0.881 & 1.612 & 0.873 & 3.847 & 1.519 & 3.726 & 1.406 \\
 & 36 & \textbf{1.583} & \textbf{0.852} & 1.715 & 0.893 & 1.702 & 0.905 & 4.115 & 1.572 & 3.931 & 1.459 \\
 & 48 & \textbf{1.638} & \textbf{0.879} & 1.803 & 0.945 & 1.855 & 0.935 & 4.243 & 1.587 & 4.069 & 1.507 \\
\cmidrule{1-12}

\multirow{4}{*}{Wind} & 12 & \textbf{0.269} & \textbf{0.263} & 0.278 & 0.297 & 0.275 & 0.279 & 0.842 & 0.567 & 0.894 & 0.475 \\
 & 24 & \textbf{0.479} & \textbf{0.399} & 0.495 & 0.432 & 0.491 & 0.426 & 1.534 & 0.759 & 1.594 & 0.723 \\
 & 36 & \textbf{0.637} & \textbf{0.488} & 0.653 & 0.514 & 0.649 & 0.518 & 2.333 & 0.922 & 2.118 & 0.883 \\
 & 48 & \textbf{0.741} & \textbf{0.549} & 0.760 & 0.568 & 0.756 & 0.585 & 3.331 & 1.082 & 2.467 & 1.002 \\
\cmidrule{1-12}

\multirow{4}{*}{Solar} & 12 & \textbf{0.115} & \textbf{0.206} & 0.124 & 0.224 & 0.122 & 0.214 & 0.768 & 0.511 & 0.136 & 0.223 \\
 & 24 & \textbf{0.216} & \textbf{0.285} & 0.230 & 0.302 & 0.227 & 0.299 & 0.877 & 0.602 & 0.258 & 0.311 \\
 & 36 & \textbf{0.288} & \textbf{0.338} & 0.300 & 0.350 & 0.297 & 0.341 & 0.985 & 0.764 & 0.341 & 0.367 \\
 & 48 & \textbf{0.276} & \textbf{0.328} & 0.292 & 0.345 & 0.288 & 0.354 & 0.915 & 0.699 & 0.329 & 0.358 \\
\bottomrule
\end{tabular}
\end{table*}

\begin{table*}[h]
\centering
\caption{Full results of the ablation study on the generative components. We compare the forecasting performance of our full framework against variants without the autoregressive mechanism (w/o AR), without the flow-matching mechanism (w/o Flow), and without both components (w/o AR-Flow) across all benchmark datasets. The best results are in \textbf{bold}.}
\label{tab:ablation_study}

\small                          
\setlength{\tabcolsep}{10pt}     
\renewcommand{\arraystretch}{1.05} 

\begin{tabular}{l|c|cc|cc|cc|cc}
\toprule

\multicolumn{2}{c|}{\textsc{Methods}} & \multicolumn{2}{c|}{\textsc{Ours}} & \multicolumn{2}{c|}{\textsc{w/o AR}} & \multicolumn{2}{c|}{\textsc{w/o Flow}} & \multicolumn{2}{c}{\textsc{w/o AR-Flow}} \\

\multicolumn{2}{c|}{\textsc{Metrics}} & MSE & MAE & MSE & MAE & MSE & MAE & MSE & MAE \\ 
\midrule


\multirow{4}{*}{Energy} & 12 & \textbf{0.140} & \textbf{0.273} & 0.184 & 0.326 & 0.178 & 0.318 & 0.187 & 0.331 \\
& 24 & \textbf{0.265} & \textbf{0.379} & 0.300 & 0.401 & 0.296 & 0.398 & 0.305 & 0.402 \\
& 36 & \textbf{0.327} & \textbf{0.428} & 0.366 & 0.445 & 0.360 & 0.442 & 0.360 & 0.441 \\
& 48 & \textbf{0.441} & \textbf{0.499} & 0.506 & 0.536 & 0.500 & 0.530 & 0.527 & 0.545 \\
\cmidrule{1-10}

\multirow{4}{*}{ETTh1} & 12 & \textbf{0.290} & \textbf{0.343} & 0.323 & 0.375 & 0.316 & 0.368 & 0.329 & 0.377 \\
& 24 & \textbf{0.319} & \textbf{0.362} & 0.358 & 0.398 & 0.350 & 0.390 & 0.364 & 0.401 \\
& 36 & \textbf{0.344} & \textbf{0.377} & 0.383 & 0.412 & 0.375 & 0.405 & 0.390 & 0.415 \\
& 48 & \textbf{0.357} & \textbf{0.386} & 0.400 & 0.423 & 0.391 & 0.416 & 0.407 & 0.426 \\
\cmidrule{1-10}

\multirow{4}{*}{ETTh2} & 12 & \textbf{0.121} & \textbf{0.216} & 0.141 & 0.234 & 0.136 & 0.231 & 0.145 & 0.239 \\ 
& 24 & \textbf{0.152} & \textbf{0.243} & 0.179 & 0.265 & 0.173 & 0.262 & 0.184 & 0.270 \\ 
& 36 & \textbf{0.172} & \textbf{0.254} & 0.200 & 0.275 & 0.194 & 0.271 & 0.205 & 0.281 \\ 
& 48 & \textbf{0.191} & \textbf{0.272} & 0.224 & 0.298 & 0.217 & 0.292 & 0.230 & 0.302 \\ 
\cmidrule{1-10}

\multirow{4}{*}{ETTm1} & 12 & \textbf{0.140} & \textbf{0.232} & 0.160 & 0.262 & 0.156 & 0.257 & 0.163 & 0.265 \\
& 24 & \textbf{0.213} & \textbf{0.280} & 0.245 & 0.318 & 0.240 & 0.312 & 0.250 & 0.322 \\
& 36 & \textbf{0.255} & \textbf{0.306} & 0.291 & 0.345 & 0.285 & 0.339 & 0.297 & 0.349 \\
& 48 & \textbf{0.288} & \textbf{0.321} & 0.331 & 0.364 & 0.324 & 0.358 & 0.338 & 0.369 \\
\cmidrule{1-10}

\multirow{4}{*}{ETTm2} & 12 & \textbf{0.079} & \textbf{0.167} & 0.096 & 0.190 & 0.093 & 0.187 & 0.101 & 0.195 \\
& 24 & \textbf{0.102} & \textbf{0.194} & 0.127 & 0.223 & 0.122 & 0.220 & 0.133 & 0.228 \\
& 36 & \textbf{0.120} & \textbf{0.212} & 0.147 & 0.242 & 0.141 & 0.238 & 0.155 & 0.247 \\
& 48 & \textbf{0.134} & \textbf{0.226} & 0.166 & 0.261 & 0.160 & 0.255 & 0.175 & 0.266 \\
\cmidrule{1-10}

\multirow{4}{*}{Environ.} & 12 & \textbf{0.286} & \textbf{0.372} & 0.325 & 0.395 & 0.320 & 0.389 & 0.332 & 0.399 \\
& 24 & \textbf{0.295} & \textbf{0.378} & 0.331 & 0.408 & 0.326 & 0.403 & 0.338 & 0.413 \\
& 36 & \textbf{0.305} & \textbf{0.383} & 0.346 & 0.415 & 0.341 & 0.410 & 0.353 & 0.420 \\
& 48 & \textbf{0.316} & \textbf{0.390} & 0.362 & 0.422 & 0.354 & 0.415 & 0.367 & 0.425 \\
\cmidrule{1-10}

\multirow{4}{*}{Exchange} & 12 & \textbf{0.015} & \textbf{0.079} & 0.028 & 0.094 & 0.025 & 0.090 & 0.030 & 0.096 \\
& 24 & \textbf{0.026} & \textbf{0.110} & 0.052 & 0.134 & 0.046 & 0.128 & 0.055 & 0.137 \\
& 36 & \textbf{0.037} & \textbf{0.132} & 0.072 & 0.158 & 0.063 & 0.151 & 0.076 & 0.162 \\
& 48 & \textbf{0.046} & \textbf{0.150} & 0.092 & 0.182 & 0.081 & 0.174 & 0.097 & 0.186 \\
\cmidrule{1-10}

\multirow{4}{*}{Health} & 12 & \textbf{1.056} & \textbf{0.697} & 1.264 & 0.745 & 1.221 & 0.740 & 1.296 & 0.754 \\
& 24 & \textbf{1.500} & \textbf{0.820} & 1.777 & 0.879 & 1.737 & 0.873 & 1.843 & 0.889 \\
& 36 & \textbf{1.583} & \textbf{0.852} & 1.973 & 0.911 & 1.831 & 0.905 & 1.943 & 0.921 \\
& 48 & \textbf{1.638} & \textbf{0.879} & 1.940 & 0.942 & 1.896 & 0.935 & 2.013 & 0.953 \\
\cmidrule{1-10}

\multirow{4}{*}{Wind} & 12 & \textbf{0.269} & \textbf{0.263} & 0.340 & 0.311 & 0.308 & 0.297 & 0.359 & 0.322 \\
& 24 & \textbf{0.479} & \textbf{0.399} & 0.608 & 0.474 & 0.552 & 0.454 & 0.642 & 0.491 \\
& 36 & \textbf{0.637} & \textbf{0.488} & 0.806 & 0.577 & 0.731 & 0.553 & 0.851 & 0.599 \\
& 48 & \textbf{0.741} & \textbf{0.549} & 0.938 & 0.642 & 0.852 & 0.624 & 0.992 & 0.676 \\
\cmidrule{1-10}

\multirow{4}{*}{Solar} & 12 & \textbf{0.115} & \textbf{0.206} & 0.143 & 0.232 & 0.136 & 0.223 & 0.149 & 0.239 \\
& 24 & \textbf{0.216} & \textbf{0.285} & 0.270 & 0.323 & 0.258 & 0.311 & 0.282 & 0.333 \\
& 36 & \textbf{0.288} & \textbf{0.338} & 0.358 & 0.381 & 0.341 & 0.367 & 0.374 & 0.393 \\
& 48 & \textbf{0.276} & \textbf{0.328} & 0.345 & 0.372 & 0.329 & 0.358 & 0.360 & 0.383 \\
\bottomrule
\end{tabular}
\end{table*}

\begin{table*}[h]
\centering
\caption{Full results of the ablation study on context features. We compare the forecasting performance of our full framework against variants without domain knowledge (w/o Domain), task instruction (w/o Instruction), statistics information (w/o Statistics), and the entire textual input (w/o Text) across all benchmark datasets. The best results are in \textbf{bold}.}
\label{tab:ablation_prompt}

\small                          
\setlength{\tabcolsep}{7pt}     
\renewcommand{\arraystretch}{1.1} 

\begin{tabular}{l|c|cc|cc|cc|cc|cc}
\toprule

\multicolumn{2}{c|}{\textsc{Methods}} & \multicolumn{2}{c|}{\textsc{Ours}} & \multicolumn{2}{c|}{\textsc{w/o domain}} & \multicolumn{2}{c|}{\textsc{w/o instruction}} & \multicolumn{2}{c|}{\textsc{w/o statistics}} & \multicolumn{2}{c}{\textsc{w/o text}} \\

\multicolumn{2}{c|}{\textsc{Metrics}} & MSE & MAE & MSE & MAE & MSE & MAE & MSE & MAE & MSE & MAE \\ 
\midrule


\multirow{4}{*}{Energy}
& 12 & \textbf{0.140} & \textbf{0.273} & 0.145 & 0.282 & 0.143 & 0.277 & 0.151 & 0.292 & 0.156 & 0.305 \\
& 24 & \textbf{0.265} & \textbf{0.379} & 0.268 & 0.381 & 0.267 & 0.380 & 0.273 & 0.383 & 0.277 & 0.386 \\
& 36 & \textbf{0.327} & \textbf{0.428} & 0.334 & 0.433 & 0.331 & 0.430 & 0.344 & 0.438 & 0.351 & 0.444 \\
& 48 & \textbf{0.441} & \textbf{0.499} & 0.445 & 0.501 & 0.443 & 0.500 & 0.452 & 0.502 & 0.457 & 0.504 \\
\cmidrule{1-12}

\multirow{4}{*}{ETTh1}
& 12 & \textbf{0.290} & \textbf{0.343} & 0.304 & 0.345 & 0.308 & 0.348 & 0.304 & 0.354 & 0.310 & 0.355 \\
& 24 & \textbf{0.319} & \textbf{0.362} & 0.330 & 0.364 & 0.333 & 0.366 & 0.330 & 0.370 & 0.335 & 0.371 \\
& 36 & \textbf{0.344} & \textbf{0.377} & 0.354 & 0.379 & 0.357 & 0.381 & 0.354 & 0.386 & 0.358 & 0.387 \\
& 48 & \textbf{0.357} & \textbf{0.386} & 0.368 & 0.388 & 0.370 & 0.390 & 0.368 & 0.395 & 0.372 & 0.396 \\
\cmidrule{1-12}

\multirow{4}{*}{ETTh2}
& 12 & \textbf{0.121} & \textbf{0.216} & 0.123 & 0.218 & 0.123 & 0.218 & 0.128 & 0.223 & 0.130 & 0.225 \\ 
& 24 & \textbf{0.152} & \textbf{0.243} & 0.154 & 0.245 & 0.155 & 0.246 & 0.161 & 0.251 & 0.164 & 0.253 \\ 
& 36 & \textbf{0.172} & \textbf{0.254} & 0.174 & 0.256 & 0.175 & 0.257 & 0.182 & 0.262 & 0.185 & 0.264 \\ 
& 48 & \textbf{0.191} & \textbf{0.272} & 0.193 & 0.274 & 0.195 & 0.275 & 0.202 & 0.281 & 0.205 & 0.283 \\ 
\cmidrule{1-12}

\multirow{4}{*}{ETTm1}
& 12 & \textbf{0.140} & \textbf{0.232} & 0.144 & 0.245 & 0.145 & 0.246 & 0.154 & 0.254 & 0.159 & 0.258 \\
& 24 & \textbf{0.213} & \textbf{0.280} & 0.216 & 0.282 & 0.217 & 0.283 & 0.224 & 0.291 & 0.228 & 0.295 \\
& 36 & \textbf{0.255} & \textbf{0.306} & 0.257 & 0.309 & 0.258 & 0.311 & 0.264 & 0.319 & 0.267 & 0.321 \\
& 48 & \textbf{0.288} & \textbf{0.321} & 0.290 & 0.323 & 0.291 & 0.325 & 0.297 & 0.332 & 0.300 & 0.335 \\
\cmidrule{1-12}

\multirow{4}{*}{ETTm2}
& 12 & \textbf{0.079} & \textbf{0.167} & 0.081 & 0.169 & 0.080 & 0.168 & 0.085 & 0.175 & 0.086 & 0.177 \\
& 24 & \textbf{0.102} & \textbf{0.194} & 0.105 & 0.196 & 0.103 & 0.195 & 0.110 & 0.203 & 0.112 & 0.206 \\
& 36 & \textbf{0.120} & \textbf{0.212} & 0.124 & 0.214 & 0.121 & 0.213 & 0.129 & 0.222 & 0.131 & 0.225 \\
& 48 & \textbf{0.134} & \textbf{0.226} & 0.138 & 0.229 & 0.136 & 0.227 & 0.144 & 0.236 & 0.147 & 0.240 \\
\cmidrule{1-12}

\multirow{4}{*}{Environ.}
& 12 & \textbf{0.286} & \textbf{0.372} & 0.288 & 0.373 & 0.287 & 0.373 & 0.290 & 0.375 & 0.291 & 0.376 \\
& 24 & \textbf{0.295} & \textbf{0.378} & 0.297 & 0.379 & 0.296 & 0.379 & 0.299 & 0.380 & 0.300 & 0.381 \\
& 36 & \textbf{0.305} & \textbf{0.383} & 0.308 & 0.384 & 0.307 & 0.384 & 0.312 & 0.386 & 0.313 & 0.387 \\
& 48 & \textbf{0.316} & \textbf{0.390} & 0.319 & 0.392 & 0.318 & 0.392 & 0.323 & 0.395 & 0.324 & 0.396 \\
\cmidrule{1-12}

\multirow{4}{*}{Exchange}
& 12 & \textbf{0.015} & \textbf{0.079} & 0.016 & 0.080 & 0.016 & 0.081 & 0.018 & 0.083 & 0.018 & 0.084 \\
& 24 & \textbf{0.026} & \textbf{0.110} & 0.027 & 0.111 & 0.027 & 0.112 & 0.029 & 0.115 & 0.029 & 0.116 \\
& 36 & \textbf{0.037} & \textbf{0.132} & 0.038 & 0.134 & 0.038 & 0.134 & 0.039 & 0.136 & 0.040 & 0.137 \\
& 48 & \textbf{0.046} & \textbf{0.150} & 0.047 & 0.153 & 0.047 & 0.153 & 0.050 & 0.158 & 0.051 & 0.160 \\
\cmidrule{1-12}

\multirow{4}{*}{Health}
& 12 & \textbf{1.056} & \textbf{0.697} & 1.063 & 0.701 & 1.074 & 0.705 & 1.090 & 0.796 & 1.112 & 0.713 \\
& 24 & \textbf{1.500} & \textbf{0.820} & 1.504 & 0.824 & 1.511 & 0.827 & 1.521 & 0.931 & 1.534 & 0.834 \\
& 36 & \textbf{1.583} & \textbf{0.852} & 1.587 & 0.855 & 1.594 & 0.857 & 1.604 & 0.962 & 1.617 & 0.862 \\
& 48 & \textbf{1.638} & \textbf{0.879} & 1.645 & 0.881 & 1.657 & 0.883 & 1.673 & 0.990 & 1.696 & 0.887 \\
\cmidrule{1-12}

\multirow{4}{*}{Wind}
& 12 & \textbf{0.269} & \textbf{0.263} & 0.272 & 0.267 & 0.271 & 0.265 & 0.275 & 0.280 & 0.277 & 0.288 \\
& 24 & \textbf{0.479} & \textbf{0.399} & 0.482 & 0.401 & 0.481 & 0.400 & 0.485 & 0.407 & 0.487 & 0.410 \\
& 36 & \textbf{0.637} & \textbf{0.488} & 0.641 & 0.490 & 0.640 & 0.489 & 0.646 & 0.495 & 0.650 & 0.498 \\
& 48 & \textbf{0.741} & \textbf{0.549} & 0.745 & 0.550 & 0.744 & 0.550 & 0.750 & 0.554 & 0.753 & 0.557 \\
\cmidrule{1-12}

\multirow{4}{*}{Solar}
& 12 & \textbf{0.115} & \textbf{0.206} & 0.117 & 0.207 & 0.119 & 0.208 & 0.120 & 0.210 & 0.122 & 0.211 \\
& 24 & \textbf{0.216} & \textbf{0.286} & 0.219 & 0.288 & 0.222 & 0.290 & 0.224 & 0.293 & 0.226 & 0.295 \\
& 36 & \textbf{0.288} & \textbf{0.338} & 0.295 & 0.345 & 0.299 & 0.347 & 0.301 & 0.351 & 0.303 & 0.354 \\
& 48 & \textbf{0.276} & \textbf{0.328} & 0.286 & 0.336 & 0.290 & 0.338 & 0.292 & 0.342 & 0.283 & 0.335 \\
\bottomrule
\end{tabular}
\end{table*}

\begin{table*}[h]
\centering
\caption{Full results of the comparative analysis among different generative methods. We compare our proposed framework with vanilla flow-matching and diffusion models across all benchmark datasets. The best results are in \textbf{bold}.}
\label{tab:generative_comparison_wide}
\setlength{\tabcolsep}{1.5pt} 
\renewcommand{\arraystretch}{1.3}
\small
\resizebox{\textwidth}{!}{%
\begin{tabular}{c c | ccc | ccc | ccc | ccc | ccc | ccc | ccc | ccc | ccc | ccc}
\toprule
\multirow{2}{*}{\textsc{Horizon}} & \multirow{2}{*}{\textsc{Metrics}} & \multicolumn{3}{c|}{\textsc{Energy}} & \multicolumn{3}{c|}{\textsc{ETTh1}} & \multicolumn{3}{c|}{\textsc{ETTh2}} & \multicolumn{3}{c|}{\textsc{ETTm1}} & \multicolumn{3}{c|}{\textsc{ETTm2}} & \multicolumn{3}{c|}{\textsc{Environ.}} & \multicolumn{3}{c|}{\textsc{Exchange}} & \multicolumn{3}{c|}{\textsc{Health}} & \multicolumn{3}{c|}{\textsc{Wind}} & \multicolumn{3}{c}{\textsc{Solar}} \\
 & & Ours & Flow & Diff & Ours & Flow & Diff & Ours & Flow & Diff & Ours & Flow & Diff & Ours & Flow & Diff & Ours & Flow & Diff & Ours & Flow & Diff & Ours & Flow & Diff & Ours & Flow & Diff & Ours & Flow & Diff \\
\midrule

\multirow{2}{*}{12} & MSE & \textbf{0.140} & 0.158 & 0.155 & \textbf{0.290} & 0.329 & 0.321 & \textbf{0.121} & 0.147 & 0.146 & \textbf{0.140} & 0.186 & 0.179 & \textbf{0.079} & 0.108 & 0.105 & \textbf{0.286} & 0.313 & 0.306 & \textbf{0.015} & 0.025 & 0.022 & \textbf{1.056} & 1.171 & 1.157 & \textbf{0.269} & 0.306 & 0.298 & \textbf{0.115} & 0.135 & 0.132 \\
 & MAE & \textbf{0.273} & 0.289 & 0.286 & \textbf{0.343} & 0.373 & 0.368 & \textbf{0.216} & 0.239 & 0.236 & \textbf{0.232} & 0.279 & 0.272 & \textbf{0.167} & 0.199 & 0.196 & \textbf{0.372} & 0.390 & 0.386 & \textbf{0.079} & 0.088 & 0.085 & \textbf{0.697} & 0.724 & 0.710 & \textbf{0.263} & 0.296 & 0.289 & \textbf{0.206} & 0.221 & 0.219 \\
\cmidrule{1-32}

\multirow{2}{*}{24} & MSE & \textbf{0.265} & 0.300 & 0.294 & \textbf{0.319} & 0.362 & 0.354 & \textbf{0.152} & 0.185 & 0.183 & \textbf{0.213} & 0.283 & 0.272 & \textbf{0.102} & 0.140 & 0.136 & \textbf{0.295} & 0.323 & 0.316 & \textbf{0.026} & 0.043 & 0.039 & \textbf{1.500} & 1.663 & 1.644 & \textbf{0.479} & 0.545 & 0.530 & \textbf{0.216} & 0.255 & 0.249 \\
 & MAE & \textbf{0.379} & 0.401 & 0.397 & \textbf{0.362} & 0.394 & 0.389 & \textbf{0.243} & 0.268 & 0.265 & \textbf{0.280} & 0.336 & 0.328 & \textbf{0.194} & 0.231 & 0.228 & \textbf{0.378} & 0.396 & 0.392 & \textbf{0.110} & 0.122 & 0.119 & \textbf{0.820} & 0.851 & 0.835 & \textbf{0.399} & 0.449 & 0.439 & \textbf{0.286} & 0.307 & 0.303 \\
\cmidrule{1-32}

\multirow{2}{*}{36} & MSE & \textbf{0.327} & 0.370 & 0.362 & \textbf{0.344} & 0.391 & 0.381 & \textbf{0.172} & 0.209 & 0.206 & \textbf{0.255} & 0.339 & 0.326 & \textbf{0.120} & 0.164 & 0.160 & \textbf{0.305} & 0.334 & 0.327 & \textbf{0.037} & 0.061 & 0.055 & \textbf{1.583} & 1.755 & 1.735 & \textbf{0.637} & 0.725 & 0.705 & \textbf{0.288} & 0.345 & 0.336 \\
 & MAE & \textbf{0.428} & 0.453 & 0.448 & \textbf{0.377} & 0.410 & 0.405 & \textbf{0.254} & 0.281 & 0.277 & \textbf{0.306} & 0.368 & 0.359 & \textbf{0.212} & 0.253 & 0.249 & \textbf{0.383} & 0.401 & 0.397 & \textbf{0.132} & 0.147 & 0.142 & \textbf{0.852} & 0.885 & 0.868 & \textbf{0.488} & 0.549 & 0.537 & \textbf{0.338} & 0.368 & 0.364 \\
\cmidrule{1-32}

\multirow{2}{*}{48} & MSE & \textbf{0.441} & 0.499 & 0.489 & \textbf{0.357} & 0.406 & 0.396 & \textbf{0.191} & 0.231 & 0.229 & \textbf{0.288} & 0.383 & 0.368 & \textbf{0.134} & 0.184 & 0.179 & \textbf{0.316} & 0.346 & 0.339 & \textbf{0.046} & 0.076 & 0.068 & \textbf{1.638} & 1.816 & 1.795 & \textbf{0.741} & 0.843 & 0.820 & \textbf{0.276} & 0.335 & 0.327 \\
 & MAE & \textbf{0.499} & 0.528 & 0.522 & \textbf{0.386} & 0.420 & 0.414 & \textbf{0.272} & 0.300 & 0.297 & \textbf{0.321} & 0.386 & 0.377 & \textbf{0.226} & 0.269 & 0.266 & \textbf{0.390} & 0.409 & 0.405 & \textbf{0.150} & 0.167 & 0.162 & \textbf{0.879} & 0.913 & 0.895 & \textbf{0.549} & 0.618 & 0.604 & \textbf{0.328} & 0.359 & 0.355 \\
\bottomrule
\end{tabular}%
}
\end{table*}

\begin{table*}[h]
\centering
\caption{Full results of the efficiency-performance trade-off analysis regarding the number of function evaluations (NFE). We compare the forecasting performance using 1, 2, and 3 sampling steps across all benchmark datasets. The best results are in \textbf{bold}.}
\label{tab:nfe_ablation_wide}
\setlength{\tabcolsep}{1.5pt} 
\renewcommand{\arraystretch}{1.3}
\small
\resizebox{\textwidth}{!}{%
\begin{tabular}{c c | ccc | ccc | ccc | ccc | ccc | ccc | ccc | ccc | ccc | ccc}
\toprule
\multirow{2}{*}{\textsc{Horizon}} & \multirow{2}{*}{\textsc{Metrics}} & \multicolumn{3}{c|}{\textsc{Energy}} & \multicolumn{3}{c|}{\textsc{ETTh1}} & \multicolumn{3}{c|}{\textsc{ETTh2}} & \multicolumn{3}{c|}{\textsc{ETTm1}} & \multicolumn{3}{c|}{\textsc{ETTm2}} & \multicolumn{3}{c|}{\textsc{Environ.}} & \multicolumn{3}{c|}{\textsc{Exchange}} & \multicolumn{3}{c|}{\textsc{Health}} & \multicolumn{3}{c|}{\textsc{Wind}} & \multicolumn{3}{c}{\textsc{Solar}} \\
 & & 1 & 2 & 3 & 1 & 2 & 3 & 1 & 2 & 3 & 1 & 2 & 3 & 1 & 2 & 3 & 1 & 2 & 3 & 1 & 2 & 3 & 1 & 2 & 3 & 1 & 2 & 3 & 1 & 2 & 3 \\
\midrule
\multirow{2}{*}{12} & MSE & 0.140 & \textbf{0.139} & 0.147 & 0.290 & \textbf{0.288} & 0.306 & 0.121 & \textbf{0.119} & 0.123 & 0.140 & \textbf{0.139} & 0.143 & 0.079 & \textbf{0.077} & 0.082 & \textbf{0.286} & \textbf{0.286} & 0.287 & \textbf{0.015} & \textbf{0.015} & 0.016 & 1.056 & \textbf{1.051} & 1.076 & 0.269 & \textbf{0.265} & 0.274 & 0.115 & \textbf{0.114} & 0.118 \\
 & MAE & 0.273 & \textbf{0.272} & 0.277 & 0.343 & \textbf{0.341} & 0.356 & 0.216 & \textbf{0.214} & 0.218 & 0.232 & \textbf{0.230} & 0.237 & 0.167 & \textbf{0.166} & 0.171 & 0.372 & \textbf{0.370} & 0.373 & 0.079 & \textbf{0.078} & 0.081 & 0.697 & \textbf{0.694} & 0.703 & 0.263 & \textbf{0.259} & 0.268 & 0.206 & \textbf{0.203} & 0.209 \\
\cmidrule{1-32}
\multirow{2}{*}{24} & MSE & 0.265 & \textbf{0.263} & 0.279 & 0.319 & \textbf{0.317} & 0.337 & 0.152 & \textbf{0.150} & 0.154 & 0.213 & \textbf{0.211} & 0.217 & 0.102 & \textbf{0.099} & 0.106 & \textbf{0.295} & \textbf{0.295} & 0.296 & 0.026 & \textbf{0.025} & 0.027 & 1.500 & \textbf{1.492} & 1.528 & 0.479 & \textbf{0.472} & 0.488 & 0.216 & \textbf{0.214} & 0.222 \\
 & MAE & 0.379 & \textbf{0.378} & 0.385 & 0.362 & \textbf{0.360} & 0.376 & 0.243 & \textbf{0.241} & 0.245 & 0.280 & \textbf{0.277} & 0.286 & 0.194 & \textbf{0.192} & 0.199 & 0.378 & \textbf{0.376} & 0.379 & 0.110 & \textbf{0.108} & 0.113 & 0.820 & \textbf{0.816} & 0.827 & 0.399 & \textbf{0.394} & 0.407 & 0.286 & \textbf{0.282} & 0.290 \\
\cmidrule{1-32}
\multirow{2}{*}{36} & MSE & 0.327 & \textbf{0.324} & 0.344 & 0.344 & \textbf{0.341} & 0.363 & 0.172 & \textbf{0.170} & 0.174 & 0.255 & \textbf{0.253} & 0.260 & 0.120 & \textbf{0.117} & 0.125 & 0.305 & \textbf{0.304} & 0.307 & 0.037 & \textbf{0.036} & 0.039 & 1.583 & \textbf{1.575} & 1.613 & 0.637 & \textbf{0.628} & 0.648 & \textbf{0.288} & 0.289 & 0.300 \\
 & MAE & 0.428 & \textbf{0.427} & 0.435 & 0.377 & \textbf{0.375} & 0.391 & 0.254 & \textbf{0.252} & 0.256 & 0.306 & \textbf{0.303} & 0.313 & 0.212 & \textbf{0.210} & 0.218 & 0.383 & \textbf{0.381} & 0.384 & 0.132 & \textbf{0.130} & 0.136 & 0.852 & \textbf{0.848} & 0.859 & 0.488 & \textbf{0.481} & 0.497 & \textbf{0.338} & \textbf{0.338} & 0.347 \\
\cmidrule{1-32}
\multirow{2}{*}{48} & MSE & 0.441 & \textbf{0.438} & 0.464 & 0.357 & \textbf{0.354} & 0.377 & 0.191 & \textbf{0.189} & 0.193 & 0.288 & \textbf{0.284} & 0.292 & 0.134 & \textbf{0.131} & 0.139 & 0.316 & \textbf{0.315} & 0.318 & 0.046 & \textbf{0.045} & 0.049 & 1.638 & \textbf{1.630} & 1.669 & 0.741 & \textbf{0.731} & 0.754 & \textbf{0.276} & 0.280 & 0.291 \\
 & MAE & 0.499 & \textbf{0.498} & 0.507 & 0.386 & \textbf{0.384} & 0.401 & 0.272 & \textbf{0.270} & 0.274 & 0.321 & \textbf{0.317} & 0.327 & 0.226 & \textbf{0.224} & 0.232 & 0.390 & \textbf{0.388} & 0.391 & 0.150 & \textbf{0.148} & 0.154 & 0.879 & \textbf{0.875} & 0.886 & 0.549 & \textbf{0.542} & 0.560 & \textbf{0.328} & 0.329 & 0.338 \\
\bottomrule
\end{tabular}%
}
\end{table*}

\begin{table*}[h]
\centering
\caption{Full results of hyperparameter analysis of the interpolation scheduler. We compare the forecasting performance using linear and cosine schedulers across all benchmark datasets. The best results are in \textbf{bold}.}
\label{tab:scheduler_ablation_wide}
\setlength{\tabcolsep}{2.5pt} 
\renewcommand{\arraystretch}{1.3}
\small
\resizebox{\textwidth}{!}{%
\begin{tabular}{c c | cc | cc | cc | cc | cc | cc | cc | cc | cc | cc}
\toprule
\multirow{2}{*}{\textsc{Horizon}} & \multirow{2}{*}{\textsc{Metrics}} & \multicolumn{2}{c|}{\textsc{Energy}} & \multicolumn{2}{c|}{\textsc{ETTh1}} & \multicolumn{2}{c|}{\textsc{ETTh2}} & \multicolumn{2}{c|}{\textsc{ETTm1}} & \multicolumn{2}{c|}{\textsc{ETTm2}} & \multicolumn{2}{c|}{\textsc{Environ.}} & \multicolumn{2}{c|}{\textsc{Exchange}} & \multicolumn{2}{c|}{\textsc{Health}} & \multicolumn{2}{c|}{\textsc{Wind}} & \multicolumn{2}{c}{\textsc{Solar}} \\
 & & Linear & Cosine & Linear & Cosine & Linear & Cosine & Linear & Cosine & Linear & Cosine & Linear & Cosine & Linear & Cosine & Linear & Cosine & Linear & Cosine & Linear & Cosine \\
\midrule
\multirow{2}{*}{12} & MSE & \textbf{0.140} & 0.141 & \textbf{0.290} & 0.293 &\textbf{0.121}&0.124& \textbf{0.140} & 0.143 & \textbf{0.079} & 0.084 & \textbf{0.286} & \textbf{0.286} & \textbf{0.015} & 0.016 & \textbf{1.056} & 1.060 & \textbf{0.269} & 0.276 & \textbf{0.115} & 0.119 \\
 & MAE & \textbf{0.273} & 0.274 & \textbf{0.343} & 0.347 &\textbf{0.216}&0.219& \textbf{0.232} & 0.237 & \textbf{0.167} & 0.173 & 0.372 & \textbf{0.370} & \textbf{0.079} & 0.081 & \textbf{0.697} & 0.705 & \textbf{0.263} & 0.266 & \textbf{0.206} & 0.210 \\
\cmidrule{1-22}
\multirow{2}{*}{24} & MSE & \textbf{0.265} & 0.267 & \textbf{0.319} & 0.322 &\textbf{0.152}&0.156& \textbf{0.213} & 0.218 & \textbf{0.102} & 0.108 & \textbf{0.295} & \textbf{0.295} & \textbf{0.026} & 0.029 & \textbf{1.500} & 1.506 & \textbf{0.479} & 0.492 & \textbf{0.216} & 0.224 \\
 & MAE & \textbf{0.379} & 0.380 & \textbf{0.362} & 0.366 &\textbf{0.243}&0.247& \textbf{0.280} & 0.286 & \textbf{0.194} & 0.201 & 0.378 & \textbf{0.376} & \textbf{0.110} & 0.112 & \textbf{0.820} & 0.829 & \textbf{0.399} & 0.403 & \textbf{0.286} & 0.292 \\
\cmidrule{1-22}
\multirow{2}{*}{36} & MSE & \textbf{0.327} & 0.329 & \textbf{0.344} & 0.348 &\textbf{0.172}&0.176& \textbf{0.255} & 0.261 & \textbf{0.120} & 0.127 & 0.305 & \textbf{0.304} & \textbf{0.037} & 0.041 & \textbf{1.583} & 1.589 & \textbf{0.637} & 0.654 & \textbf{0.288} & 0.297 \\
 & MAE & \textbf{0.428} & 0.429 & \textbf{0.377} & 0.381 &\textbf{0.254}&0.258& \textbf{0.306} & 0.313 & \textbf{0.212} & 0.220 & 0.383 & \textbf{0.381} & \textbf{0.132} & 0.135 & \textbf{0.852} & 0.861 & \textbf{0.488} & 0.493 & \textbf{0.338} & 0.345 \\
\cmidrule{1-22}
\multirow{2}{*}{48} & MSE & \textbf{0.441} & 0.444 & \textbf{0.357} & 0.361 &\textbf{0.191}&0.196& \textbf{0.288} & 0.293 & \textbf{0.134} & 0.142 & 0.316 & \textbf{0.315} & \textbf{0.046} & 0.050 & \textbf{1.638} & 1.645 & \textbf{0.741} & 0.761 & \textbf{0.276} & 0.285 \\
 & MAE & \textbf{0.499} & 0.501 & \textbf{0.386} & 0.390 &\textbf{0.272}&0.276& \textbf{0.321} & 0.327 & \textbf{0.226} & 0.234 & 0.390 & \textbf{0.388} & \textbf{0.150} & 0.153 & \textbf{0.879} & 0.889 & \textbf{0.549} & 0.554 & \textbf{0.328} & 0.335 \\
\bottomrule
\end{tabular}%
}
\end{table*}

\begin{table*}[h]
\centering
\caption{Full results of hyperparameter analysis of the patch size. We evaluate the impact of different patch sizes $\{2, 4, 6, 8, 10\}$ on forecasting performance across all benchmark datasets. The best results are in \textbf{bold}.}
\label{tab:patchsize_ablation}

\small                        
\setlength{\tabcolsep}{8pt}     
\renewcommand{\arraystretch}{1.05} 

\begin{tabular}{l|c|cc|cc|cc|cc|cc}
\toprule

\multicolumn{2}{c|}{\textsc{PatchSize}} & \multicolumn{2}{c|}{ 2} & \multicolumn{2}{c|}{4} & \multicolumn{2}{c|}{6} & \multicolumn{2}{c|}{8} & \multicolumn{2}{c}{ 10} \\

\multicolumn{2}{c|}{\textsc{Metrics}} & MSE & MAE & MSE & MAE & MSE & MAE & MSE & MAE & MSE & MAE \\ 
\midrule


\multirow{4}{*}{Energy}
& 12 & 0.144 & 0.278 & \textbf{0.140} & \textbf{0.273} & 0.147 & 0.283 & 0.151 & 0.289 & 0.153 & 0.291 \\
& 24 & 0.272 & 0.386 & \textbf{0.265} & \textbf{0.379} & 0.278 & 0.393 & 0.286 & 0.401 & 0.290 & 0.404 \\
& 36 & 0.336 & 0.436 & \textbf{0.327} & \textbf{0.428} & 0.343 & 0.443 & 0.353 & 0.453 & 0.358 & 0.456 \\
& 48 & 0.453 & 0.508 & \textbf{0.441} & \textbf{0.499} & 0.463 & 0.517 & 0.477 & 0.528 & 0.483 & 0.532 \\
\cmidrule{1-12}

\multirow{4}{*}{ETTh1}
& 12 & 0.319 & 0.370 & 0.296 & 0.351 & \textbf{0.290} & \textbf{0.343} & 0.339 & 0.389 & 0.299 & 0.354 \\
& 24 & 0.362 & 0.396 & 0.333 & 0.375 & \textbf{0.319} & \textbf{0.362} & 0.385 & 0.417 & 0.340 & 0.379 \\
& 36 & 0.400 & 0.417 & 0.365 & 0.394 & \textbf{0.344} & \textbf{0.377} & 0.425 & 0.439 & 0.375 & 0.399 \\
& 48 & 0.420 & 0.429 & 0.382 & 0.404 & \textbf{0.357} & \textbf{0.386} & 0.447 & 0.451 & 0.394 & 0.410 \\
\cmidrule{1-12}

\multirow{4}{*}{ETTh2}
& 12 & 0.157 & 0.242 & \textbf{0.121} & \textbf{0.216} & 0.140 & 0.234 & 0.137 & 0.228 & 0.140 & 0.222 \\
& 24 & 0.197 & 0.273 & \textbf{0.152} & \textbf{0.243} & 0.175 & 0.263 & 0.171 & 0.256 & 0.175 & 0.249 \\
& 36 & 0.222 & 0.285 & \textbf{0.172} & \textbf{0.254} & 0.198 & 0.275 & 0.194 & 0.268 & 0.198 & 0.261 \\
& 48 & 0.246 & 0.305 & \textbf{0.191} & \textbf{0.272} & 0.220 & 0.294 & 0.215 & 0.287 & 0.220 & 0.279 \\
\cmidrule{1-12}

\multirow{4}{*}{ETTm1}
& 12 & 0.144 & 0.238 & \textbf{0.140} & \textbf{0.232} & 0.154 & 0.245 & 0.155 & 0.246 & 0.152 & 0.244 \\
& 24 & 0.220 & 0.287 & \textbf{0.213} & \textbf{0.280} & 0.234 & 0.296 & 0.236 & 0.297 & 0.231 & 0.294 \\
& 36 & 0.263 & 0.314 & \textbf{0.255} & \textbf{0.306} & 0.280 & 0.323 & 0.282 & 0.325 & 0.277 & 0.321 \\
& 48 & 0.297 & 0.329 & \textbf{0.288} & \textbf{0.321} & 0.316 & 0.339 & 0.319 & 0.340 & 0.312 & 0.337 \\
\cmidrule{1-12}

\multirow{4}{*}{ETTm2}
& 12 & 0.097 & 0.188 & 0.094 & 0.185 & \textbf{0.079} & \textbf{0.167} & 0.081 & 0.171 & 0.093 & 0.183 \\
& 24 & 0.126 & 0.219 & 0.122 & 0.215 & \textbf{0.102} & \textbf{0.194} & 0.105 & 0.198 & 0.121 & 0.213 \\
& 36 & 0.148 & 0.239 & 0.143 & 0.235 & \textbf{0.120} & \textbf{0.212} & 0.124 & 0.217 & 0.142 & 0.233 \\
& 48 & 0.165 & 0.255 & 0.160 & 0.250 & \textbf{0.134} & \textbf{0.226} & 0.138 & 0.231 & 0.158 & 0.248 \\
\cmidrule{1-12}

\multirow{4}{*}{Environ.}
& 12 & \textbf{0.286} & \textbf{0.372} & 0.305 & 0.397 & 0.315 & 0.408 & 0.315 & 0.409 & 0.441 & 0.481 \\
& 24 & \textbf{0.295} & \textbf{0.378} & 0.316 & 0.406 & 0.330 & 0.417 & 0.332 & 0.421 & 0.437 & 0.481 \\
& 36 & \textbf{0.305} & \textbf{0.383} & 0.331 & 0.416 & 0.346 & 0.427 & 0.355 & 0.436 & 0.445 & 0.485 \\
& 48 & \textbf{0.316} & \textbf{0.390} & 0.343 & 0.425 & 0.363 & 0.438 & 0.379 & 0.453 & 0.451 & 0.488 \\
\cmidrule{1-12}

\multirow{4}{*}{Exchange}
& 12 & \textbf{0.015} & 0.081 & \textbf{0.015} & \textbf{0.079} & \textbf{0.015} & 0.080 & \textbf{0.015} & \textbf{0.079} & \textbf{0.015} & 0.080 \\
& 24 & 0.028 & 0.113 & \textbf{0.026} & 0.110 & \textbf{0.026} & \textbf{0.109} & \textbf{0.026} & \textbf{0.109} & \textbf{0.026} & \textbf{0.109} \\
& 36 & 0.040 & 0.137 & \textbf{0.037} & 0.132 & \textbf{0.037} & \textbf{0.131} & \textbf{0.037} & 0.132 & \textbf{0.037} & \textbf{0.131} \\
& 48 & 0.053 & 0.158 & \textbf{0.046} & \textbf{0.150} & 0.048 & \textbf{0.150} & 0.049 & 0.153 & 0.048 & \textbf{0.150} \\
\cmidrule{1-12}

\multirow{4}{*}{Health}
& 12 & 1.759 & 0.940 & 1.327 & 0.788 & \textbf{1.056} & \textbf{0.697} & 2.966 & 1.298 & 2.831 & 1.258 \\
& 24 & 1.984 & 1.011 & 1.750 & 0.909 & \textbf{1.500} & \textbf{0.820} & 2.946 & 1.290 & 2.755 & 1.245 \\
& 36 & 2.058 & 1.021 & 1.757 & 0.918 & \textbf{1.583} & \textbf{0.852} & 2.900 & 1.270 & 2.755 & 1.229 \\
& 48 & 2.126 & 1.052 & 1.912 & 0.965 & \textbf{1.638} & \textbf{0.879} & 2.974 & 1.278 & 2.788 & 1.235 \\
\cmidrule{1-12}

\multirow{4}{*}{Wind}
& 12 & 0.298 & 0.295 & \textbf{0.269} & \textbf{0.263} & 0.277 & 0.271 & 0.289 & 0.287 & 0.302 & 0.299 \\
& 24 & 0.530 & 0.448 & \textbf{0.479} & \textbf{0.399} & 0.493 & 0.411 & 0.515 & 0.435 & 0.537 & 0.454 \\
& 36 & 0.705 & 0.548 & \textbf{0.637} & \textbf{0.488} & 0.656 & 0.502 & 0.684 & 0.532 & 0.714 & 0.555 \\
& 48 & 0.820 & 0.617 & \textbf{0.741} & \textbf{0.549} & 0.763 & 0.565 & 0.796 & 0.598 & 0.831 & 0.624 \\
\cmidrule{1-12}

\multirow{4}{*}{Solar}
& 12 & 0.122 & 0.215 & \textbf{0.115} & \textbf{0.206} & 0.121 & 0.213 & 0.124 & 0.217 & 0.130 & 0.225 \\
& 24 & 0.230 & 0.298 & \textbf{0.216} & \textbf{0.286} & 0.228 & 0.295 & 0.233 & 0.301 & 0.244 & 0.311 \\
& 36 & 0.311 & 0.358 & \textbf{0.288} & \textbf{0.338} & 0.309 & 0.354 & 0.315 & 0.361 & 0.331 & 0.374 \\
& 48 & 0.302 & 0.349 & \textbf{0.276} & \textbf{0.328} & 0.300 & 0.345 & 0.306 & 0.352 & 0.322 & 0.365 \\
\bottomrule
\end{tabular}
\end{table*}

\begin{table*}[h]
\centering
\caption{Full results of the comparative analysis on backbone architectures. We evaluate the performance of the vanilla transformer against  Qwen LLM family with varying parameter scales (0.6B, 1.7B, and 4B) across all benchmark datasets. The best results are in \textbf{bold}.}
\label{tab:scaling_analysis}

\small                          
\setlength{\tabcolsep}{10pt}     
\renewcommand{\arraystretch}{1} 

\begin{tabular}{l|c|cc|cc|cc|cc}
\toprule

\multicolumn{2}{c|}{\textsc{Backbone}} & \multicolumn{2}{c|}{\textsc{Transformer}} & \multicolumn{2}{c|}{\textsc{Qwen3-0.6B}} & \multicolumn{2}{c|}{\textsc{Qwen3-1.7B}} & \multicolumn{2}{c}{\textsc{Qwen3-4B}} \\

\multicolumn{2}{c|}{\textsc{Metrics}} & MSE & MAE & MSE & MAE & MSE & MAE & MSE & MAE \\ 
\midrule


\multirow{4}{*}{Energy}
& 12 & 0.144 & 0.276 & 0.140 & 0.273 & 0.141 & 0.274 & \textbf{0.138} & \textbf{0.271} \\
& 24 & 0.272 & 0.383 & 0.265 & 0.379 & 0.267 & 0.380 & \textbf{0.262} & \textbf{0.376} \\
& 36 & 0.336 & 0.433 & 0.327 & 0.428 & 0.329 & 0.429 & \textbf{0.323} & \textbf{0.425} \\
& 48 & 0.453 & 0.504 & 0.441 & 0.499 & 0.444 & 0.501 & \textbf{0.436} & \textbf{0.495} \\
\cmidrule{1-10}

\multirow{4}{*}{ETTh1}
& 12 & 0.305 & 0.356 & 0.290 & 0.343 & 0.290 & 0.341 & \textbf{0.286} & \textbf{0.339} \\
& 24 & 0.335 & 0.376 & 0.319 & 0.362 & 0.319 & 0.360 & \textbf{0.315} & \textbf{0.358} \\
& 36 & 0.361 & 0.391 & 0.344 & 0.377 & 0.344 & 0.375 & \textbf{0.339} & \textbf{0.373} \\
& 48 & 0.375 & 0.401 & 0.357 & 0.386 & 0.356 & 0.384 & \textbf{0.352} & \textbf{0.382} \\
\cmidrule{1-10}

\multirow{4}{*}{ETTh2}
& 12 & 0.131 & 0.226 & 0.121 & 0.216 & 0.122 & 0.217 & \textbf{0.118} & \textbf{0.211} \\
& 24 & 0.164 & 0.255 & 0.152 & 0.243 & 0.153 & 0.244 & \textbf{0.148} & \textbf{0.238} \\
& 36 & 0.186 & 0.266 & 0.172 & 0.254 & 0.173 & 0.255 & \textbf{0.168} & \textbf{0.249} \\
& 48 & 0.206 & 0.285 & 0.191 & 0.272 & 0.192 & 0.273 & \textbf{0.186} & \textbf{0.266} \\
\cmidrule{1-10}

\multirow{4}{*}{ETTm1}
& 12 & 0.151 & 0.242 & 0.140 & 0.232 & 0.139 & 0.231 & \textbf{0.137} & \textbf{0.229} \\
& 24 & 0.229 & 0.292 & 0.213 & 0.280 & 0.212 & 0.279 & \textbf{0.208} & \textbf{0.276} \\
& 36 & 0.274 & 0.319 & 0.255 & 0.306 & 0.254 & 0.305 & \textbf{0.249} & \textbf{0.302} \\
& 48 & 0.309 & 0.334 & 0.288 & 0.320 & 0.286 & 0.319 & \textbf{0.281} & \textbf{0.316} \\
\cmidrule{1-10}

\multirow{4}{*}{ETTm2}
& 12 & 0.092 & 0.182 & 0.079 & 0.167 & 0.079 & 0.169 & \textbf{0.076} & \textbf{0.165} \\
& 24 & 0.118 & 0.212 & 0.102 & 0.194 & 0.102 & 0.196 & \textbf{0.098} & \textbf{0.191} \\
& 36 & 0.139 & 0.231 & 0.120 & 0.212 & 0.120 & 0.214 & \textbf{0.115} & \textbf{0.209} \\
& 48 & 0.155 & 0.247 & 0.134 & 0.226 & 0.134 & 0.229 & \textbf{0.128} & \textbf{0.223} \\
\cmidrule{1-10}

\multirow{4}{*}{Environ.}
& 12 & 0.296 & 0.379 & 0.286 & 0.372 & 0.285 & 0.371 & \textbf{0.283} & \textbf{0.367}\\
& 24 & 0.305 & 0.385 & 0.295 & 0.378 & 0.294 & 0.377 & \textbf{0.292} & \textbf{0.373} \\
& 36 & 0.316 & 0.390 & 0.305 & 0.383 & 0.304 & 0.382 & \textbf{0.301} & \textbf{0.378} \\
& 48 & 0.327 & 0.397 & 0.316 & 0.390 & 0.316 & 0.389 & \textbf{0.312} & \textbf{0.385} \\
\cmidrule{1-10}

\multirow{4}{*}{Exchange}
& 12 & 0.018 & 0.083 & 0.015 & 0.079 & 0.015 & 0.079 & \textbf{0.014} & \textbf{0.077} \\
& 24 & 0.031 & 0.115 & 0.026 & 0.110 & 0.026 & 0.109 & \textbf{0.024} & \textbf{0.107} \\
& 36 & 0.044 & 0.138 & 0.037 & 0.132 & 0.037 & 0.131 & \textbf{0.035} & \textbf{0.129} \\
& 48 & 0.055 & 0.157 & 0.046 & 0.150 & 0.046 & 0.149 & \textbf{0.043} & \textbf{0.146} \\
\cmidrule{1-10}

\multirow{4}{*}{Health}
& 12 & 1.105 & 0.706 & 1.056 & 0.697 & 1.056 & 0.696 & \textbf{1.052} & \textbf{0.694} \\
& 24 & 1.570 & 0.831 & 1.500 & 0.820 & 1.500 & 0.819 & \textbf{1.494} & \textbf{0.816} \\
& 36 & 1.657 & 0.863 & 1.583 & 0.852 & 1.583 & 0.851 & \textbf{1.577} & \textbf{0.848} \\
& 48 & 1.715 & 0.891 & 1.638 & 0.879 & 1.638 & 0.878 & \textbf{1.632} & \textbf{0.875} \\
\cmidrule{1-10}

\multirow{4}{*}{Wind}
& 12 & 0.297 & 0.285 & 0.269 & 0.263 & 0.268 & 0.263 & \textbf{0.266} & \textbf{0.258} \\
& 24 & 0.529 & 0.433 & 0.479 & 0.399 & 0.478 & 0.398 & \textbf{0.473} & \textbf{0.392} \\
& 36 & 0.704 & 0.530 & 0.637 & 0.488 & 0.635 & 0.487 & \textbf{0.629} & \textbf{0.479} \\
& 48 & 0.818 & 0.596 & 0.741 & 0.549 & 0.739 & 0.548 & \textbf{0.732} & \textbf{0.539} \\
\cmidrule{1-10}

\multirow{4}{*}{Solar}
& 12 & 0.137 & 0.222 & 0.115 & 0.206 & 0.116 & 0.206 & \textbf{0.114} & \textbf{0.203} \\
& 24 & 0.258 & 0.308 & 0.216 & 0.286 & 0.218 & 0.286 & \textbf{0.214} & \textbf{0.282} \\
& 36 & 0.347 & 0.370 & \textbf{0.288} & \textbf{0.338} & 0.295 & 0.343 & 0.289 & \textbf{0.338} \\
& 48 & 0.338 & 0.360 & \textbf{0.276} & \textbf{0.328} & 0.285 & 0.333 & 0.280 & \textbf{0.328} \\
\bottomrule
\end{tabular}
\end{table*}

\begin{table*}[h]
\centering
\caption{Full results of the  analysis on the look-back window length. We evaluate the impact of extended look-back window length $L \in \{96, 192, 336\}$ on forecasting performance across all benchmark datasets. The best results are in \textbf{bold}.}
\label{tab:look_back_ablation_wide}
\setlength{\tabcolsep}{1.5pt} 
\renewcommand{\arraystretch}{1.2}
\small
\resizebox{\textwidth}{!}{%
\begin{tabular}{c c | ccc | ccc | ccc | ccc | ccc | ccc | ccc | ccc | ccc | ccc}
\toprule
\multirow{2}{*}{\textsc{Horizon}} & \multirow{2}{*}{\textsc{Metrics}} & \multicolumn{3}{c|}{\textsc{Energy}} & \multicolumn{3}{c|}{\textsc{ETTh1}} & \multicolumn{3}{c|}{\textsc{ETTh2}} & \multicolumn{3}{c|}{\textsc{ETTm1}} & \multicolumn{3}{c|}{\textsc{ETTm2}} & \multicolumn{3}{c|}{\textsc{Environ.}} & \multicolumn{3}{c|}{\textsc{Exchange}} & \multicolumn{3}{c|}{\textsc{Health}} & \multicolumn{3}{c|}{\textsc{Wind}} & \multicolumn{3}{c}{\textsc{Solar}} \\
 & & 96 & 192 & 336 & 96 & 192 & 336 & 96 & 192 & 336 & 96 & 192 & 336 & 96 & 192 & 336 & 96 & 192 & 336 & 96 & 192 & 336 & 96 & 192 & 336 & 96 & 192 & 336 & 96 & 192 & 336 \\
\midrule
\multirow{2}{*}{12} & MSE & 0.140 & 0.138 & \textbf{0.137} & 0.290 & 0.289 & \textbf{0.284} & 0.121 & 0.120 & \textbf{0.118} & 0.140 & 0.139 & \textbf{0.137} & 0.079 & 0.078 & \textbf{0.077} & 0.286 & 0.285 & \textbf{0.266} & 0.015 & 0.015 & \textbf{0.014} & 1.056 & 1.052 & \textbf{1.044} & 0.269 & 0.267 & \textbf{0.260} & 0.115 & 0.115 & \textbf{0.113} \\
 & MAE & 0.273 & 0.272 & \textbf{0.270} & 0.343 & 0.342 & \textbf{0.338} & 0.216 & 0.214 & \textbf{0.211} & 0.232 & 0.231 & \textbf{0.229} & 0.167 & 0.168 & \textbf{0.166} & 0.372 & 0.371 & \textbf{0.366} & 0.079 & 0.081 & \textbf{0.077} & 0.697 & 0.695 & \textbf{0.691} & 0.263 & 0.261 & \textbf{0.256} & 0.206 & 0.207 & \textbf{0.202} \\
\cmidrule{1-32}
\multirow{2}{*}{24} & MSE & 0.265 & 0.262 & \textbf{0.260} & 0.319 & 0.318 & \textbf{0.313} & 0.152 & 0.151 & \textbf{0.148} & 0.213 & 0.212 & \textbf{0.208} & 0.102 & 0.101 & \textbf{0.099} & 0.295 & 0.294 & \textbf{0.275} & 0.026 & 0.027 & \textbf{0.024} & 1.500 & 1.495 & \textbf{1.483} & 0.479 & 0.475 & \textbf{0.463} & 0.216 & 0.216 & \textbf{0.213} \\
 & MAE & 0.379 & 0.377 & \textbf{0.374} & 0.362 & 0.361 & \textbf{0.357} & 0.243 & 0.241 & \textbf{0.238} & 0.280 & 0.279 & \textbf{0.276} & 0.194 & 0.195 & \textbf{0.192} & 0.378 & 0.377 & \textbf{0.372} & 0.110 & 0.112 & \textbf{0.107} & 0.820 & 0.818 & \textbf{0.813} & 0.399 & 0.395 & \textbf{0.388} & 0.286 & 0.287 & \textbf{0.281} \\
\cmidrule{1-32}
\multirow{2}{*}{36} & MSE & 0.327 & 0.323 & \textbf{0.321} & 0.344 & 0.342 & \textbf{0.337} & 0.172 & 0.171 & \textbf{0.168} & 0.255 & 0.254 & \textbf{0.249} & 0.120 & 0.119 & \textbf{0.117} & 0.305 & 0.303 & \textbf{0.284} & 0.037 & 0.038 & \textbf{0.035} & 1.583 & 1.577 & \textbf{1.565} & 0.637 & 0.632 & \textbf{0.616} & \textbf{0.288} & 0.292 & \textbf{0.288} \\
 & MAE & 0.428 & 0.426 & \textbf{0.423} & 0.377 & 0.376 & \textbf{0.372} & 0.254 & 0.252 & \textbf{0.249} & 0.306 & 0.305 & \textbf{0.302} & 0.212 & 0.213 & \textbf{0.210} & 0.383 & 0.382 & \textbf{0.377} & 0.132 & 0.135 & \textbf{0.129} & 0.852 & 0.850 & \textbf{0.845} & 0.488 & 0.484 & \textbf{0.475} & 0.338 & 0.344 & \textbf{0.337} \\
\cmidrule{1-32}
\multirow{2}{*}{48} & MSE & 0.441 & 0.436 & \textbf{0.433} & 0.357 & 0.355 & \textbf{0.350} & 0.191 & 0.190 & \textbf{0.186} & 0.288 & 0.286 & \textbf{0.281} & 0.134 & 0.133 & \textbf{0.131} & 0.316 & 0.314 & \textbf{0.294} & 0.046 & 0.047 & \textbf{0.043} & 1.638 & 1.632 & \textbf{1.620} & 0.741 & 0.735 & \textbf{0.717} & \textbf{0.276} & 0.283 & 0.279 \\
 & MAE & 0.499 & 0.497 & \textbf{0.493} & 0.386 & 0.385 & \textbf{0.381} & 0.272 & 0.270 & \textbf{0.266} & 0.321 & 0.319 & \textbf{0.316} & 0.226 & 0.227 & \textbf{0.224} & 0.390 & 0.389 & \textbf{0.384} & 0.150 & 0.153 & \textbf{0.146} & 0.879 & 0.877 & \textbf{0.871} & 0.549 & 0.544 & \textbf{0.534} & \textbf{0.328} & 0.335 & \textbf{0.328} \\
\bottomrule
\end{tabular}%
}
\end{table*}


\end{document}